\documentclass[runningheads]{llncs}
\usepackage{eccv}
\usepackage{eccvabbrv}

\usepackage{graphicx}
\usepackage{booktabs}
\usepackage[dvipsnames]{xcolor}
\usepackage[accsupp]{axessibility}
\usepackage{bbm}
\usepackage{array}
\usepackage{makecell}
\usepackage{tabularray}
\usepackage{multirow}

\usepackage[pagebackref,breaklinks,colorlinks,citecolor=eccvblue]{hyperref}
\usepackage{orcidlink}

\title{Invisible Stitch: Generating Smooth 3D Scenes with Depth Inpainting} 
\titlerunning{Invisible Stitch}
\author{Paul Engstler \and
Andrea Vedaldi \and
Iro Laina \and
Christian Rupprecht
}
\authorrunning{P.~Engstler et al.}

\institute{Visual Geometry Group, University of Oxford
\email{\{paule,vedaldi,iro,chrisr\}@robots.ox.ac.uk}\\
\url{https://research.paulengstler.com/invisible-stitch}}

\begin{document}
\maketitle
\begin{abstract}
3D scene generation has quickly become a challenging new research direction, fueled by consistent improvements of 2D generative diffusion models.
Most prior work in this area generates scenes by iteratively stitching newly generated frames with existing geometry. 
These works often depend on pre-trained monocular depth estimators to lift the generated images into 3D, fusing them with the existing scene representation.
These approaches are then often evaluated via a text metric, measuring the similarity between the generated images and a given text prompt.
In this work, we make two fundamental contributions to the field of 3D scene generation.
First, we note that lifting images to 3D with a monocular depth estimation model is suboptimal as it ignores the geometry of the existing scene. We thus introduce a novel depth completion model, trained via teacher distillation and self-training to learn the 3D fusion process, resulting in improved geometric coherence of the scene.
Second, we introduce a new benchmarking scheme for scene generation methods that is based on ground truth geometry, and thus measures the quality of the structure of the scene.
\end{abstract}
\section{Introduction}%
\label{sec:intro}

The rise of high-quality image generative models is opening up many new interesting computer vision applications.
Especially the field of novel-view synthesis is being transformed through leveraging the visual priors learned by large-scale generative models.
In this field, a new direction is emerging: scene generation. 
Here, the goal is to generate not just a new image or view, but a whole scene, often starting from a single input image or text description.
To this end, current approaches apply an iterative process of alternating geometry estimation, moving the camera and inpainting holes until the whole scene is generated.
To estimate the geometry of the scene, these methods rely on general-purpose depth estimation models, that can infer relative or absolute scene depth from a single input image. 
This approach, however, often results in inconsistencies, as the depth prediction is purely based on image information and does not take into account the geometry of the already existing scene.

Currently, scene generation is evaluated only visually, through image-based metrics such as the CLIP score between a scene description and individual views. 
This evaluation does not assess the geometry of the scene and often simply evaluates the image quality of the generator instead of evaluating the generated scene.

Thus, in this paper, we make two contributions to the field of scene generation. 
Firstly, we introduce a rigorous scene geometry evaluation benchmark, based on real and synthetic scene datasets.
The idea is to use ground truth images and depth and evaluate the geometry of generated scenes on the basis of depth maps.
Given a view of a scene, a method is tasked to generate a scene from a given view point, for which there exists a ground-truth depth map. 
The generated scene geometry can then be easily evaluated. 
Secondly, we introduce a new general-purpose depth estimation model that can be conditioned on an incomplete depth map.
Incomplete depth maps are obtained by projecting the existing scene into a new view point, where occlusions and unseen parts of the scene leave holes. 
In this manner, the newly predicted depth is conditional on the previous generations and results in a greatly increased consistency across the scene.

Our conditional depth prediction model can be trained in a simple self-supervised way.
We fine-tune an existing depth prediction model and condition it on partially masked depth maps.
This can be done on a simple image dataset using pseudo-ground-truth from the model that is being fine-tuned and does not rely on camera poses or other annotations.
To obtain realistic depth masks, we automatically create a dataset of masks by warping predicted depth maps to random views.
This allows masking depth maps with masks similar to the ones that will be encountered during inference when generating scenes. 

In our experiments, we show that existing scene generation methods suffer from geometric inconsistencies that are uncovered by our new benchmark. 
Furthermore, we can show that our conditional depth prediction model drastically reduces these artefacts as it, by design, is trained to retain geometric consistency across subsequent frames.
Our method obtains state-of-the-art performance and is general: it can be used in other scene-generation approaches for improved consistency.

We summarize the contributions of our work as follows:
\begin{itemize}
    \item We introduce a depth-inpainting model that extrapolates depth in 3D scene generation tasks.
    \item We provide a new benchmark for evaluating the geometric quality of scene generation methods. 
    \item We show that our approach generates depth that is consistent with the existing scene and is superior to prior alignment approaches that utilize unconditioned monocular depth estimation.
\end{itemize}
\section{Related Work}%
\label{s:related}

\paragraph{Depth Completion}
With the emergence of depth sensing technologies, inferring a dense depth map of a 3D scene from a sparse depth representation and a given RGB image has gained significant importance. Works in this field seek to integrate cues from both modalities either in a 2D \cite{cheng2018depth, cheng2020cspn, park2020non, zhang2023completionformer} or 3D feature space \cite{chen2019learning, huynh2021boosting, wang2018deep, boulch2020fkaconv, kam2022costdcnet} to produce a complete depth map.

While these methods are able to recover the depth of an entire scene from a possibly very sparse depth input, they have not been designed to complete depth for regions without any depth information, which naturally occur in a scene generation task.

\paragraph{3D Scene Generation}
Text-to-3D or image-to-3D scene generation has seen tremendous progress in recent years, where the majority of works in this field can either be categorized as object-centric \cite{seitz97photorealistic, yu2021pixelnerf, jain2021putting, deng2022depth, trevithick21grf:, sitzmann19deepvoxels:, lin2023magic3d, liu2023zero, melas2023realfusion, poole2022dreamfusion, wang2024prolificdreamer, qian2023magic123, shi2023mvdream, raj2023dreambooth3d}, i.e., focusing on objects without background, or holistic, generating a single 3D scene or 3D trajectories with background.

Earlier methods in the former group are only concerned with novel view synthesis, not considering the scene's geometry. These have been based on layer-structured representations  \cite{tulsiani18layer-structured, mildenhall19local, li21mine:, shih203d-photography}, e.g., layered depth images, or more implicit ones, such as in SynSin~\cite{wiles20synsin:}. More moderns approaches typically distill a 3D representation, such as a NeRF~\cite{mildenhall2021nerf} or 3D Gaussians~\cite{kerbl20233d}, from the supervision of a 2D image generation model like Stable Diffusion~\cite{rombach2022high}. Other works directly learn a 3D representation from 2D images \cite{chan2021pi, chan2022efficient, gu2021stylenerf, nguyen2019hologan, niemeyer2021giraffe, szymanowicz2023splatter}.

More holistic methods consider generating entire scenes beyond a single object. These methods generally build a scene in a sequential manner using supervision from 2D image generation models. PixelSynth~\cite{rockwell21pixelsynth:}, Text2Room~\cite{hollein23text2room:}, Text2NeRF~\cite{zhang23text2nerf:}, LucidDreamer~\cite{chung23luciddreamer:}, WonderJourney~\cite{yu23wonderjourney:}, Infinite Nature~\cite{liu2021infinite} and Text2Immersion~\cite{ouyang23text2immersion:} rely on an off-the-shelf general-purpose depth estimation model, such as ZoeDepth~\cite{bhat2023zoedepth}, to project the hallucinated 2D scene extensions into a 3D representation. The Denoising Diffusion Vision Model~\cite{saxena2023surprising} has been proposed as another backbone for this approach.
Other approaches learn an implicit representation, such as 
GAUDI~\cite{bautista22gaudi:}, ZeroNVS \cite{sargent2023zeronvs}, DiffDreamer~\cite{cai23diffdreamer:}, and InfiniteNature-Zero~\cite{li2022infinitenature}. LDM3D \cite{stan2023ldm3d}, 3D-aware Image Generation using 2D Diffusion Models~\cite{xiang233d-aware}, and RGBD$^{2}$~\cite{lei23rgbd2:} fuse image and depth prediction to generate a scene. More specialized methods introduce different representations, such as BlockFusion~\cite{wu24blockfusion:}, Worldsheet~\cite{hu21worldsheet:}, and Set-the-Scene~\cite{cohen-bar23set-the-scene:}.

While multiple works explicitly acknowledge the importance of consistent scene geometry, it has not been considered broadly. As we consider scene generation an inherently geometric task, we dedicate it our full attention in this work.

\section{Method}%
\label{sec:method}

Current 3D scene generation methods rely on 2D-based models like Stable Diffusion \cite{rombach2022high} to hallucinate scenes beyond known regions and lift generated images into three dimensions by utilizing depth estimation networks.
In this task, it is crucial to \emph{seamlessly} integrate the newly hallucinated regions into the existing scene representation.
Previous methods \cite{hollein23text2room:, zhang23text2nerf:, ouyang23text2immersion:, chung23luciddreamer:, yu23wonderjourney:, liu2021infinite} align these by applying simple global scale-and-shift operations to the predicted depth map along with other minor modifications, if any. These operations, however, might be too coarse and insufficiently local, potentially causing the transition between the scene and its hallucinated extension to be not perfectly smooth, leading to visible discontinuities.

\subsection{3D Scene Generation}

\begin{figure}
    \centering
    \includegraphics[width=0.9\textwidth]{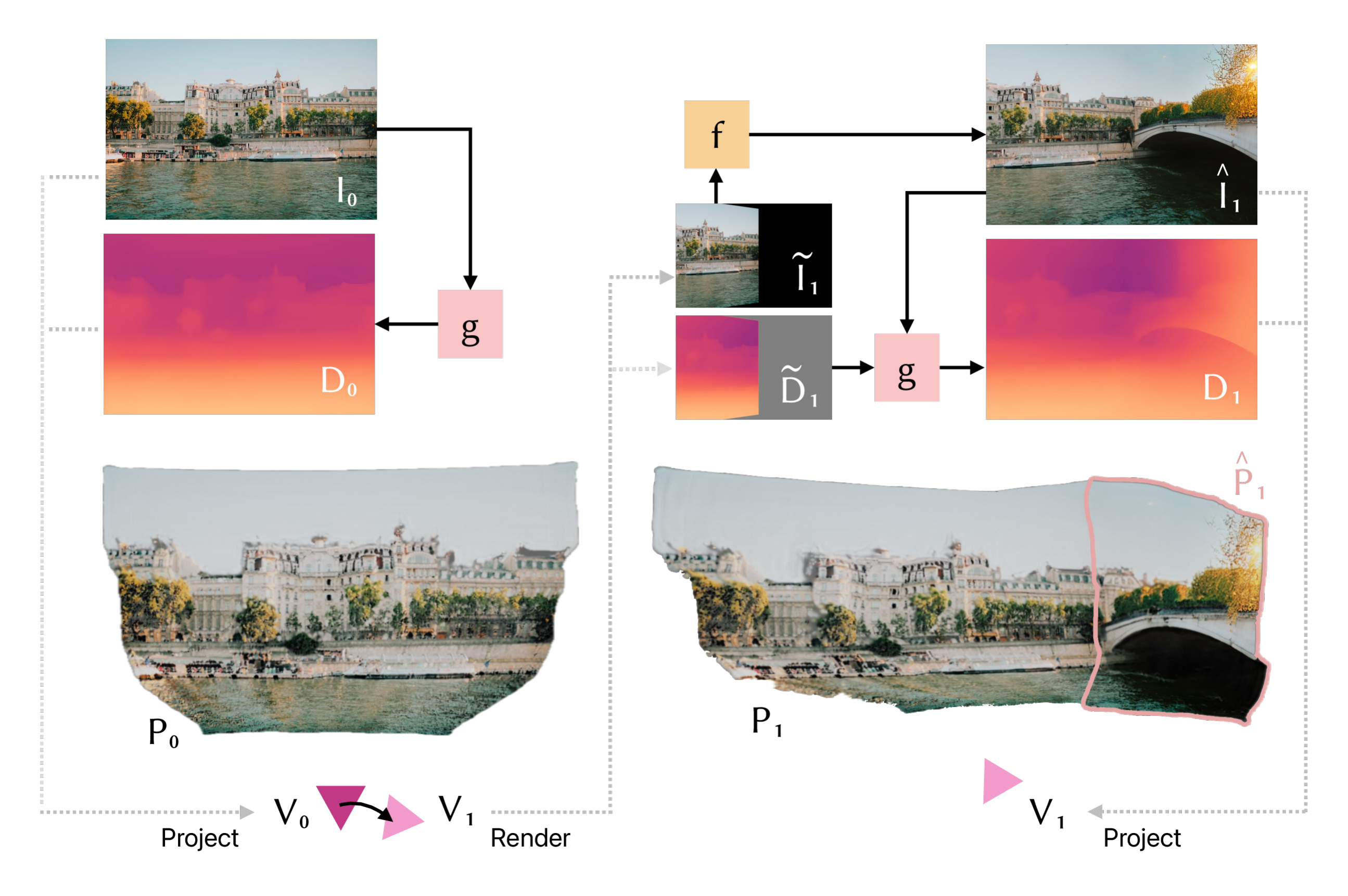}
    \caption{\textbf{Overview of our 3D scene generation method.} Starting from an input image $I_0$, we project it to a point cloud based on a depth map predicted by a depth estimation network $g$. To extend the scene, we render it from a new view point and query a generative model $f$ to hallucinate beyond the scene's boundary. Now, we condition $g$ on the depth of the existing scene and the image of the scene extended by $f$ to produce a geometrically consistent depth map to project the hallucinated points. This process may be repeated until a 360-degree scene has been generated.}
    \label{fig:projection}
\end{figure}

The task of scene generation from a single image $I_0 \in \mathbb{R}^{3 \times \Omega}$ (where $\Omega = \{1, \ldots, H\}\times\{1, \ldots, W \}$ is a lattice representing pixels) can be formulated as follows.
Given an arbitrary view point $V = [R \vert t] \in \textbf{SE}(3)$ and an intrinsic camera matrix $K$, the task is to provide a view of the scene $\hat{I}(V, K) \in \mathbb{R}^{3 \times \Omega}$ that is consistent with the original image and any other views that have already been generated.
To achieve this consistency, we parametrise the scene as a pointcloud $\mathcal{P} = \{ (C_j, X_j) \}_j$ of points at 3D locations $X_j \in \mathbb{R}^3$ with color $C_j \in \mathbb{R}^3$.

Generating a new view from a pointcloud can be done by projecting the 3D points to pixels into the image plane of the new view $x_j \equiv KV^{-1}X_j$ (since $x_j \in \mathbb{R}^2$, here $\equiv$ represents the mapping from homogeneous coordinates to image coordinates).
However, this forward projection can leave holes in the resulting image $\tilde{I}(x_i) = C_j$ when the camera is looking at previously unseen regions. We thus also obtain a binary mask $M = \{0, 1\}^\Omega$ that indicates these holes in the image.

To complete the sparse projection $\tilde{I}$ one can leverage a large-scale generative model $f$ (\eg, Stable Diffusion~\cite{rombach2022high}) which has learned a visual prior for a massive collection of visual data.
In particular, we use an inpainting variant of the Stable Diffusion model that has been trained to fill in missing regions in an image $\hat{I} = f(\tilde{I}, M)$.

In 3D scene generation from a single image or a text description, one does not have access to a 3D pointcloud of the scene. 
Instead, the goal is to generate said pointcloud.
We obtain $\mathcal{P}$ through an iterative process that by design enforces the consistency between each view of the scene. A natural mapping from images to pointclouds can be established via depth maps $D \in \mathbb{R}^{H \times W}$ as it allows the projection of the image pixels into the scene 

\begin{equation}
    \hat{\mathcal{P}}_i =  \left\{ V_i K_i^{-1} \begin{bmatrix} 
        u \\ v \\ D_i(u,v)
    \end{bmatrix} 
    \right\}_{(u,v) \in \Omega} .
\end{equation}

We construct $\mathcal{P}$ by iteratively expanding the scene representation. 
Let $\mathcal{P}_i$ be the pointcloud at the $i$-th iteration. 
Each iteration expands the representation with new geometry as $\mathcal{P}_{i+1} = \mathcal{P}_i \cup \hat{\mathcal{P}}_{i+1}$.

At each iteration, we choose a new view point $V_i$ and camera matrix $K_i$ to expand the scene.
We achieve this by passing the current image $\hat{I}_i$, which corresponds to a new generation, to a depth estimation network to obtain the corresponding depth map $D_i$.
Specifically, the depth estimation network is defined as $D_i = g(\hat{I}_i, M_i, \tilde{D}_i)$, \ie, it takes as additional inputs the mask $M_i$, which signifies which pixels have no depth estimate (since $\hat{I}_i$ is an inpainted image), and $\tilde{D}_i$, which is the depth map obtained by projecting the pointcloud $\mathcal{P}_{i-1}$ into the current view (and thus contains holes).
Note that since the depth map $\tilde{D}_i$ is obtained by projecting the existing point cloud $\mathcal{P}_{i-1}$ into the current view, only the holes indicated by $M_i$ that get filled in by $g$ contribute new points to $\mathcal{P}_i$. The other points already exist in the scene.

The depth prediction model $g$ plays a critical role, as it uses a complete image and a partial depth map to predict new scene geometry that matches the image and the existing geometry.
Prior work uses off-the-shelf depth prediction models, that are not conditional on existing geometry. They thus need to rely on heuristics such as depth interpolation to make the new prediction consistent with the scene which introduces strong artifacts. 

In the next section, we will describe our training scheme for learning the depth completion model $g$.

\subsection{Unsupervised Depth Completion}

\begin{figure}
    \centering
    \includegraphics[width=0.85\textwidth]{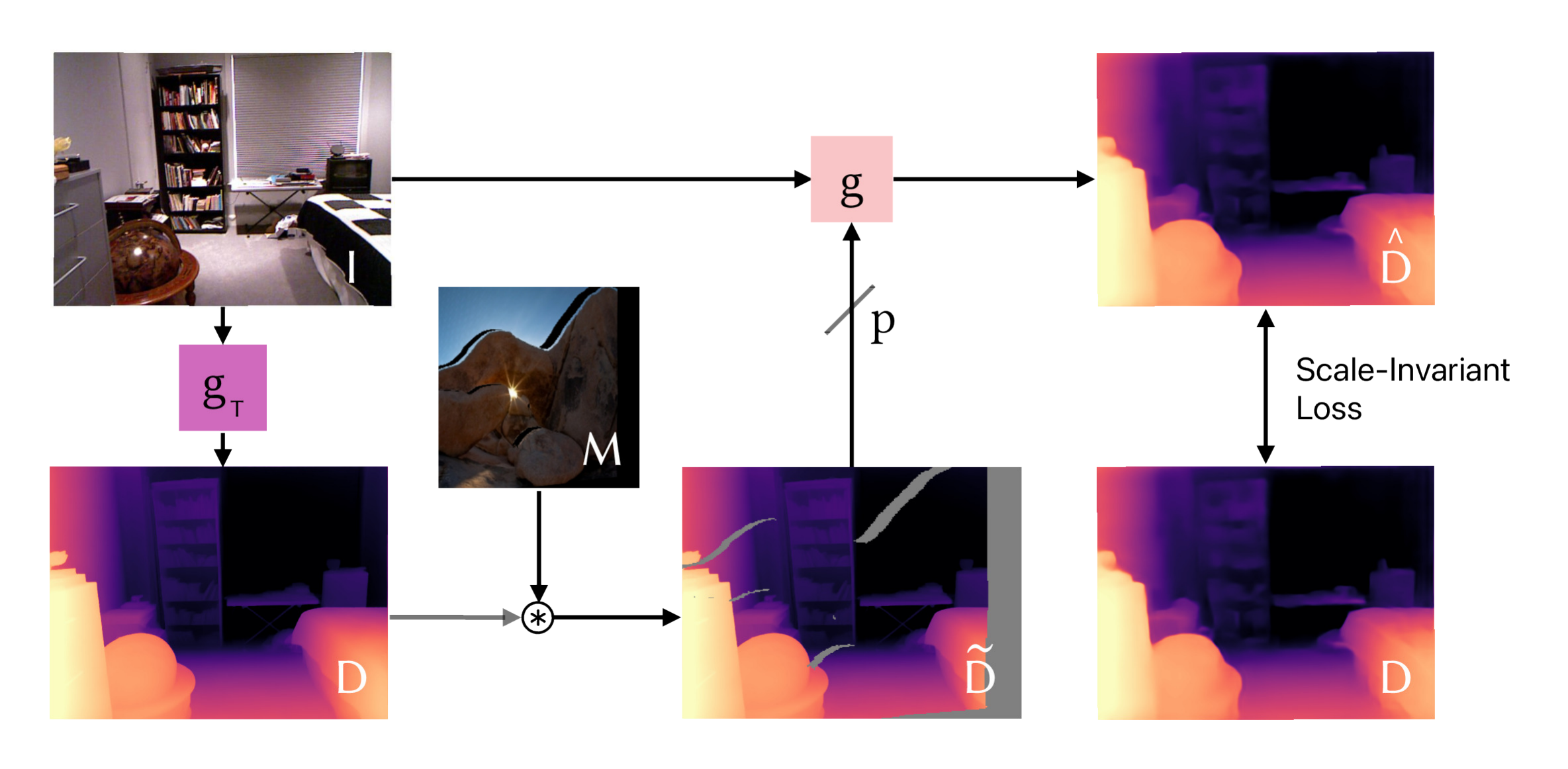}
    \caption{\textbf{Overview of our training procedure.} In this compact training scheme a depth completion network $g$ is learned by jointly training depth inpainting as well as depth prediction without a sparse depth input (the ratio being determined by the task probability $p$). A teacher network $g_T$ is utilized to generate a pseudo ground-truth depth map $D$ for a given image $I$. This depth map is then masked with a random mask $M$, to obtain a sparse depth input $\tilde{D}$.}
    \label{fig:training}
\end{figure}

Our goal is to learn a model $g(I, M, \tilde{D})$ for the purpose of 3D scene generation that provides a robust depth estimate given an image $\hat{I}$, a depth-mask $M$, and a partial depth map $\tilde{D}$.

Naturally, this task can be trained with supervision, given a multi-view dataset with known camera poses and depth ground truth.
However, these datasets are usually comparatively small and limited to specific scene types (\eg, indoors, driving, etc.).
Yet scene generation is a broad task and models should generalise well to any kind of scene. 
We thus train $g$ in a self-supervised fashion using an off-the-shelf general-purpose monocular depth prediction model $g_T(I)$ that predicts unconditional depth from a single RGB image. 
This model takes on the role of the teacher model in a student-teacher training scheme. 

Given a dataset of only images $I_k$, we generate a pseudo-labelled training dataset for $g$ as follows.
For each image in the dataset, we obtain a target depth map from a teacher $g_T$, $D_k = g_T(I_k)$.
Then, for each depth map, and similar to the scene generation step, we sample one or more random view points $V_l$ and camera matrix $K_l$ and we warp the depth map $D_k$ to the new view point obtaining a mask $M_{k,l}$ and reprojected depth $D_{k,l}$.
We collect all masks generated this way in a set $\mathcal{M} = \{ M_{k,l}  \}_{k,l}$ that represents the typical occlusion patterns generated by view point changes.

Now, given the lack of multi-view data, during training, we sample a random mask from $M_n \in \mathcal{M}$ ($1 \leq n \leq |\mathcal{M}|$) for each image $I_k$ in the batch. And train $g$ to reconstruct the pseudo depth $D_k$ guided by the scale-invariant loss \cite{eigen2014depth}, where $\tilde{d} = g(I_k, M_n, D_k \odot M_n)$, $d = D_k$, and $\psi_i = log \: \tilde{d_i} - log \: d_i$.

\begin{equation}
    \mathcal{L}_\mathrm{depth} = \sqrt{
    \frac{1}{T}
    \sum_i \psi_i^2 - \frac{\lambda}{T^2}(\sum_i \psi_i)^2
    }
\end{equation}

T is the number of pixels in $D_k$ that have valid ground-truth values. This scheme allows learning $g$ only from pseudo-supervision. 
While a na{\"i}ve scheme would use random masks, instead of masks generated by warping depth maps, we found that a model trained this way does not generalize well to the particular patterns intrinsic to depth projection.

An additional benefit is that $g$ can be initialized with a depth estimation model itself, effectively fine-tuning it for depth completion.
Moreover, we can then retain its original depth prediction (instead of depth completion) capabilities by choosing $M_n = 0$ with probability $p$, effectively masking all input depth, and recovering the depth prediction task.
Finally, as is typical (but not necessary) in student-teacher training, we can choose $g_T$ as a large model while $g$ can be a more lightweight architecture, which improves $g$ via distillation.
\section{Scene Geometry Evaluation Benchmark}
\label{sec:sgeb}
Within the fully generative task of scene generation, evaluating the geometric properties of generated scenes is difficult due to the lack of ground-truth data.
As a result, most existing work resorts to image-text similarity scores, such as the CLIP-Score~\cite{hessel2021clipscore}, which only measures the global semantic alignment of the generation with a text description.
This leaves open questions about the geometric consistency and quality of the depth predictions used to build the scene.
To answer these questions, we propose a new evaluation benchmark that quantifies the depth-reconstruction quality on a partial scene with known ground truth depth. More specifically, we seek to measure the deviation between the extrapolated depth and the ground-truth.

\begin{figure}
    \centering
    \includegraphics[width=\textwidth]{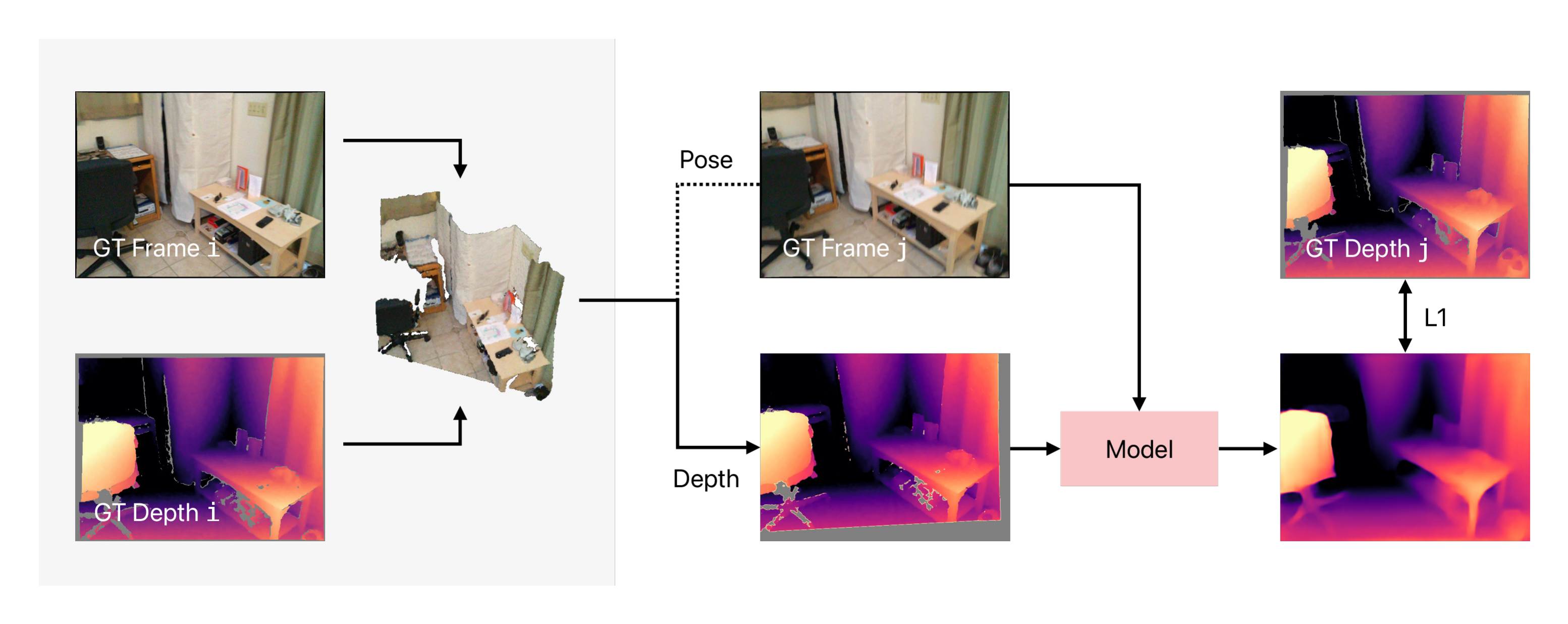}
    \caption{\textbf{Overview of our scene consistency evaluation approach.} Assume a scene is described by a set of views $\{v_1, v_2, \dots\}$ with associated images, depth maps, and camera poses, where the overlap of two views is described by a function $\phi(v_i, v_j)$. For a given view pair $(v_i, v_j)$ with $\phi(v_i, v_j) \geq \tau$, we generate a representation, e.g., a point cloud, from the ground-truth (GT) data for $v_i$. Then, we render the representation from the view point of $v_j$. We feed the corresponding ground-truth image and the representation's depth into the model under consideration to extrapolate the missing depth. Finally, we calculate the mean absolute error between the result and the ground-truth depth for $v_j$, only considering those regions that were extrapolated.}
    \label{fig:sce}
\end{figure}

\subsubsection{Approach}
Starting from a scene representation fully constructed from the ground-truth information of one view, we seek to extrapolate the depth for another ground-truth view that is highly overlapping with the first one. As the depth is known for the second view, we compute the error between the generated depth and the ground-truth depth. We only consider the error in regions that were extrapolated.
A detailed description of this approach is provided in \Cref{fig:sce}.

We use a point cloud as our representation of choice and base the overlap of two views $\phi(v_i, v_j)$ on the number of pixels that show part of the $v_i$ scene from the view point of $v_j$. Put differently, if a rendering pipeline renders an image with dimensions $H \times W$ and assigns a default value $x$ for a pixel $p$ that does not represent any parts of a scene, we define $\phi$ as:
\begin{equation}
   \phi(v_i, v_j) = \frac{\sum_{i, j}^{H \times W} \mathbbm{1}_{p(i, j) \neq x}}{H \times W}
\end{equation}

\subsubsection{Datasets}
In our evaluation, we consider ScanNet~\cite{dai2017scannet} and Hypersim~\cite{roberts2021hypersim} as they provide images, dense depth, and camera poses to accurately reconstruct scenes. They boast a wide range of scenes with varying complexity, making them an ideal test bed for evaluating the quality of depth predictions. As the former is a real-world dataset featuring indoor scenes and the latter is a photorealistic one, they lie within the distributions of most depth estimation models.

Scenes in ScanNet are described by highly-overlapping sequential frames. Thus, we chunk them into blocks of 50 frames and consider the first and tenth frame in each block for our evaluation. This allows us to yield ample views from each scene and maintains diversity, while limiting the number of evaluations to run. As the sequential frames are naturally highly overlapping, we refrain from setting a specific threshold $\tau$. To maintain reasonable evaluation times, we only consider the first 50 scenes, which yields a total of 7,832 view pairs.

With Hypersim, we compute $\phi$ across all views of a single camera trajectory within a scene and set $\tau := 0.8$.
We exclude scenes rendered with non-standard projection matrices\footnote{See \url{https://github.com/apple/ml-hypersim/issues/24}}.
The resulting number of view pairs that we evaluate on is 19,243.

For both datasets, we report the average absolute error on the extrapolated region across all pairs of views across all scenes.

\section{Experiments}

\subsection{Implementation Details}
We fine-tune a pre-trained ZoeDepth model to obtain our depth completion model $g$, re-initializing its patch embedding layer to receive two additional channels apart from the image input. These channels provide the sparse depth input $\tilde{D}$ as well as a mask $M$ describing the presence of sparse depth, i.e., $\tilde{D} > 0$. While we only replace this layer, we keep the entire model unfrozen to ensure the additional information can be integrated in later layers. We set $\lambda = 0.85$ in the scale-invariant loss.

We train on images from the NYU~Depth~v2~\cite{Silberman:ECCV12} dataset, using the monocular depth estimation network Marigold~\cite{ke2023repurposing} as a teacher $g_T$ to distill its prediction capabilities into our depth estimation network $g$.

To construct the set of warped masks $\mathcal{M}$, which contains typical masking patterns seen with view point changes, we consider the Places365~\cite{zhou2017places} dataset, generating one mask from each image. Masks are randomly chosen to be applied to a training sample depth $D$.

Ensuring we retain the original depth prediction task, we set the probability to zero out the sparse depth input to $p := 0.5$.

\subsection{Generating 360-Degree Scenes}
\label{ssec:generating-360deg-scenes}
Having shown that our model produces depth with properties beneficial to scene generation, we now utilize it for this very task. Specifically, we aim to generate immersive 360-degree scenes starting from a single real-world image. We will first outline the design of our pipeline to generate the scene, and then present quantitative and qualitative results.

\subsubsection{Pipeline Design}
Our pipeline relies on multiple components to generate a scene from a single image: First, in the same vein as current 3D scene generation methods, we enlist the help of a Stable Diffusion inpainting model ($f$) to hallucinate how a scene looks like beyond its boundaries. Second, we use our depth inpainting model ($g$) to produce an initial depth estimation for the original image and inpaint the depth in subsequent steps to attach the extrapolations. Third, we utilize Gaussian splat optimization~\cite{kerbl20233d} to obtain a smooth representation, filling in the gaps between points.

\paragraph{Generating the Point Cloud}
Starting from a given single image, we obtain a depth estimation to project it to a point cloud. We use a stationary perspective camera with fixed intrinsics. With each step, we rotate the camera slightly further along its azimuth to obtain a view that provides a canvas for the Stable Diffusion model to inpaint while still partially including the existing scene.

When inpainting images with Stable Diffusion, distortion artifacts have been known to appear in those regions that are not supposed to be edited, which has been attributed to its variational autoencoder \cite{zhu2023designing}. These alterations then cause a mismatch between the input image and the sparse depth input, which is difficult to resolve. To minimize these effects, we utilize an asymmetric autoencoder \cite{zhu2023designing} that emphasizes the decoder, making it heavier than the encoder and providing it with additional information about the inpainting task. We find that this autoencoder leads to a significant decrease in the prevalence of these artifacts.

Once the expanded scene has been visually hallucinated by Stable Diffusion, we pass the image onto our depth inpainting model with the depth of the existing scene. We project all hallucinated pixels based on the depth prediction, which seamlessly connect to the point cloud without the need for any alignment steps. We observe that depth predictions might have a gradient instead of a hard boundary at object edges, which leads to floaters radiating around objects in the point cloud. To minimize their occurrence, we identify regions in the predicted depth map with a high gradient, mask them, and assign them new values based on their nearest neighbors. This \textit{snaps} pixels in these gradient regions either to the object or its surrounding, creating a hard boundary.

We repeat this process until the loop is closed, yielding a 360-degree scene. We make sure that the final hallucination step has a wide canvas to connect both ends of the loop, assuming there has been a slight domain shift between the original image and the cascade of hallucinated views.

\paragraph{Adding Support Views}
While the scene is complete from the view of the stationary camera, it might have lots of holes due to occlusions. To \textit{fill out} the scene, we generate additional views where we employ a look-at camera that looks \textit{behind} objects, enabling us to hallucinate disocclusion regions.

For a given view from the point cloud generation phase, we determine that view's center point by averaging its depth values and define a camera looking at that particular point, with small random azimuth and elevation rotation values. As this camera might uncover unobserved regions of a scene, we query Stable Diffusion and our depth inpainting model again to extend the point cloud.

Additionally, these views also further constrain the Gaussian splat optimization process, yielding better results due to less ambiguity.

\paragraph{Gaussian Splatting} As the final step of our pipeline, we turn the obtained point cloud into Gaussians that we optimize with the hallucinated and supporting views that we obtained from the previous two phases.

We add this step to our pipeline to move away from the limiting point cloud representation to achieve a smooth scene, filling additional holes that might be present in this scene.

\subsubsection{Qualitative Results}

\begin{figure}
    \centering
    \begin{tabular}{c @{\hspace{5\tabcolsep}} cc @{\hspace{5\tabcolsep}} cc @{\hspace{5\tabcolsep}} cc}
     Prompt & \multicolumn{2}{c}{\makecell{Prague during \\ the golden hour}} & \multicolumn{2}{c}{\makecell{A street with \\ traditional buildings \\ in Kyoto, Japan}} & \multicolumn{2}{c}{\makecell{A suburban street \\ in North Carolina on \\ a bright, sunny day}} \\[0.5cm]
        Input & \includegraphics[width=0.14\textwidth]{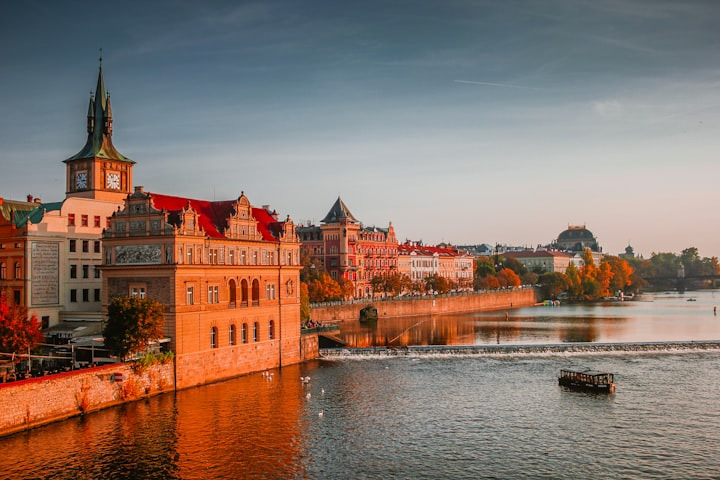} & \includegraphics[width=0.14\textwidth]{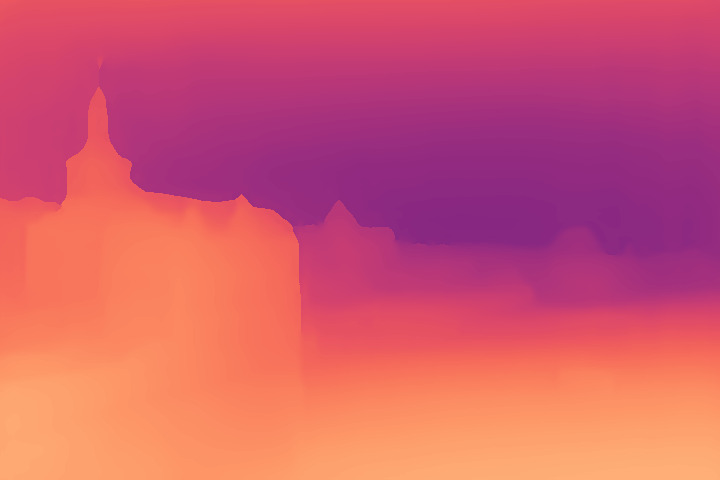} & \includegraphics[width=0.14\textwidth]{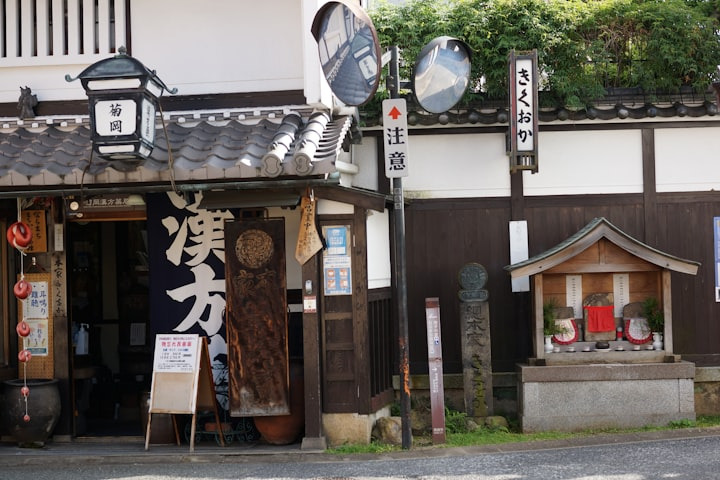} & \includegraphics[width=0.14\textwidth]{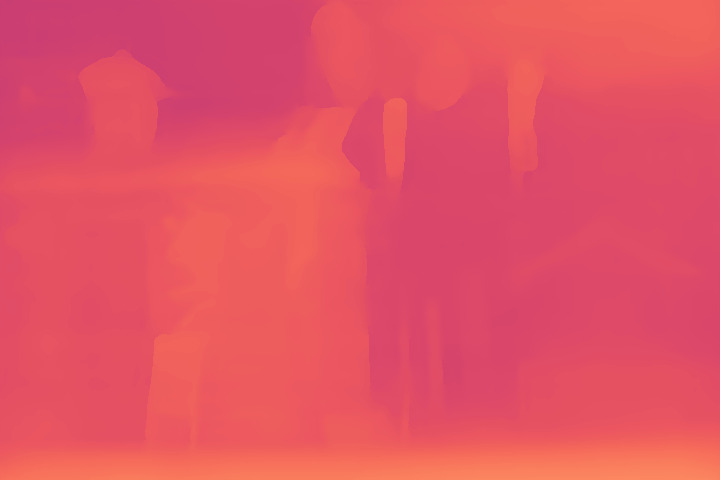} & \includegraphics[width=0.14\textwidth]{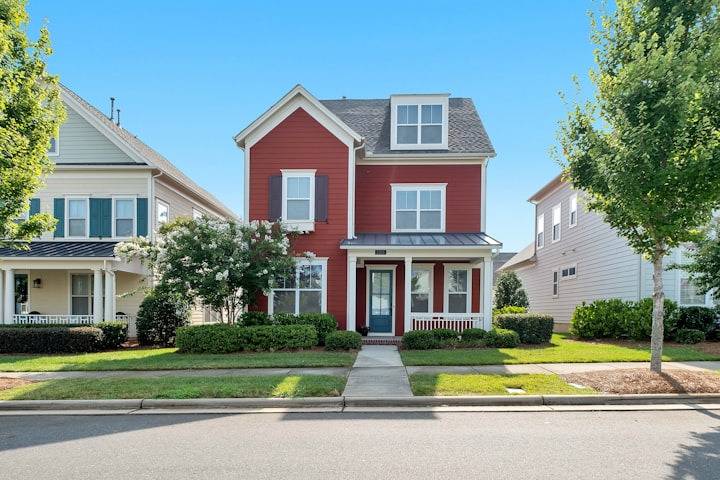} & \includegraphics[width=0.14\textwidth]{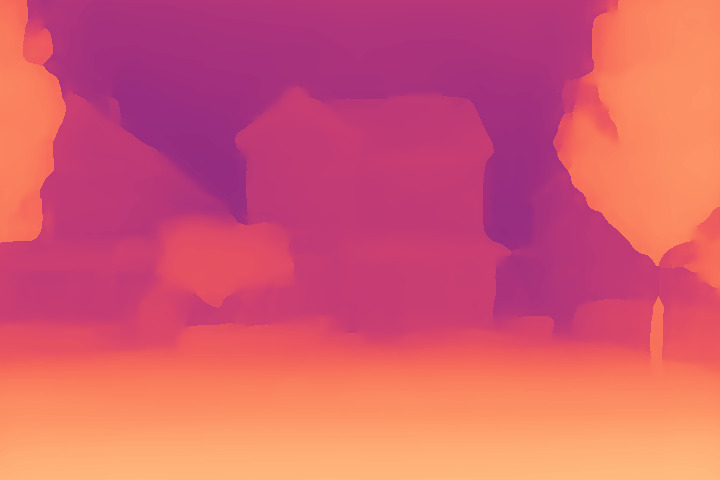} \\
        25$^{\circ}$ & \includegraphics[width=0.14\textwidth]{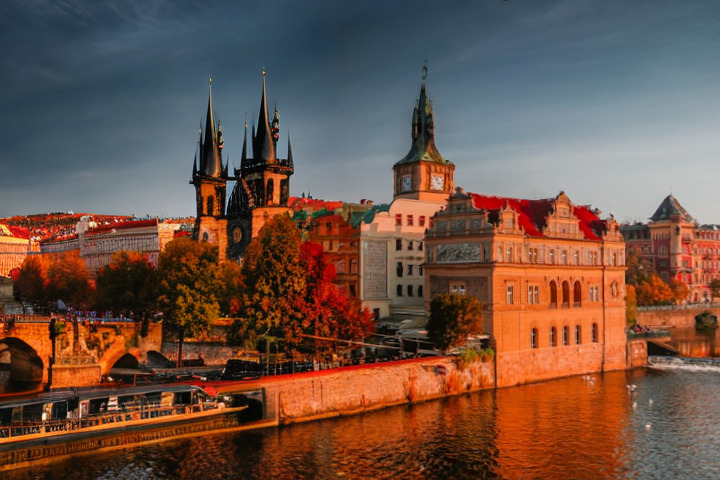} & \includegraphics[width=0.14\textwidth]{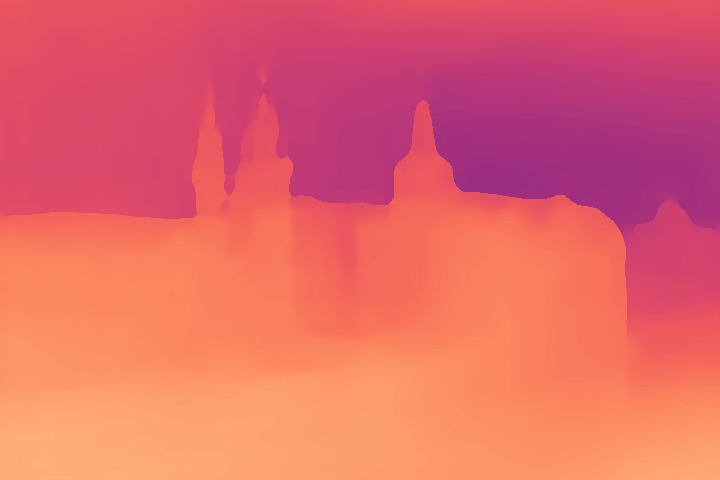} & \includegraphics[width=0.14\textwidth]{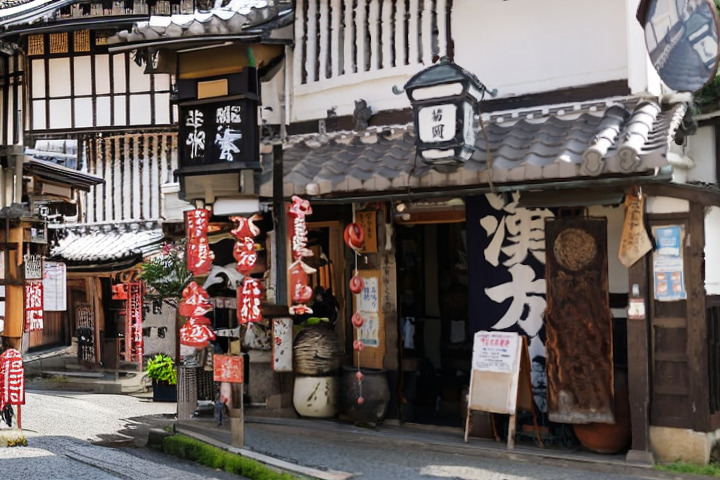} & \includegraphics[width=0.14\textwidth]{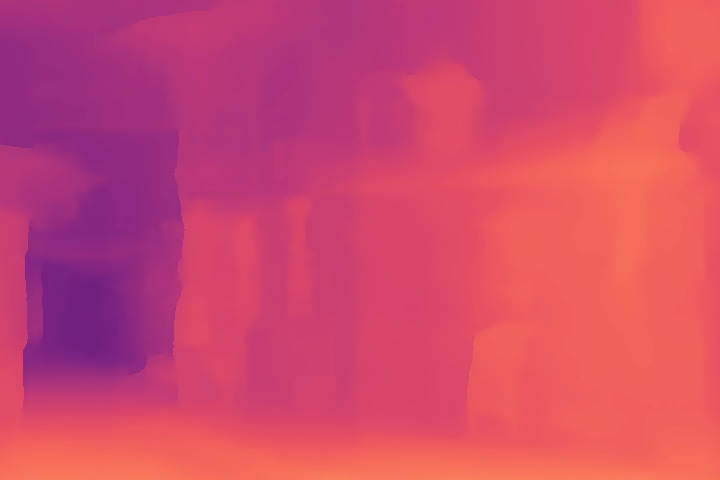} &
        \includegraphics[width=0.14\textwidth]{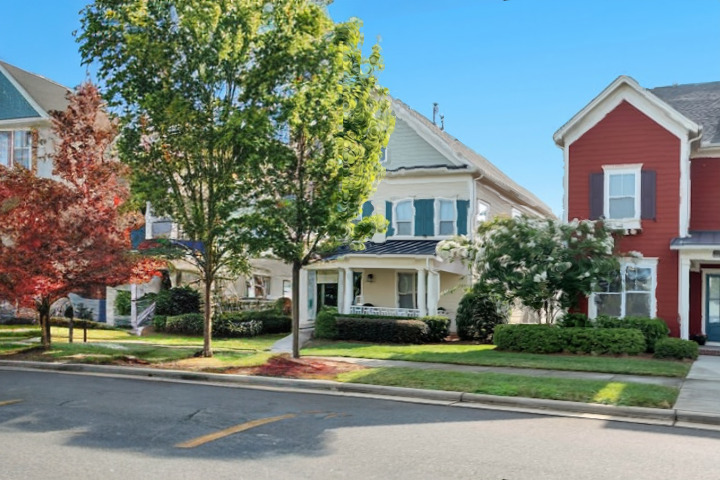} & \includegraphics[width=0.14\textwidth]{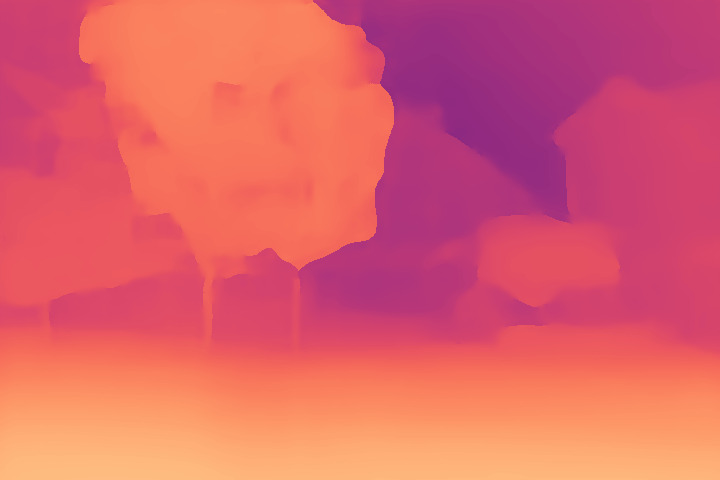} \\
        50$^{\circ}$ & \includegraphics[width=0.14\textwidth]{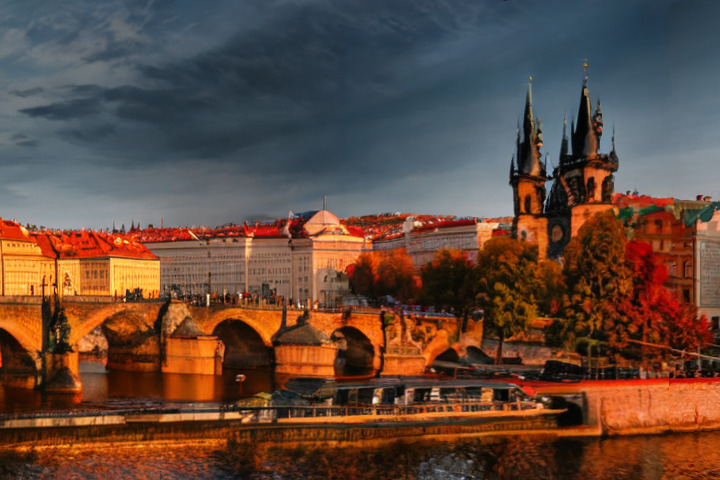} & \includegraphics[width=0.14\textwidth]{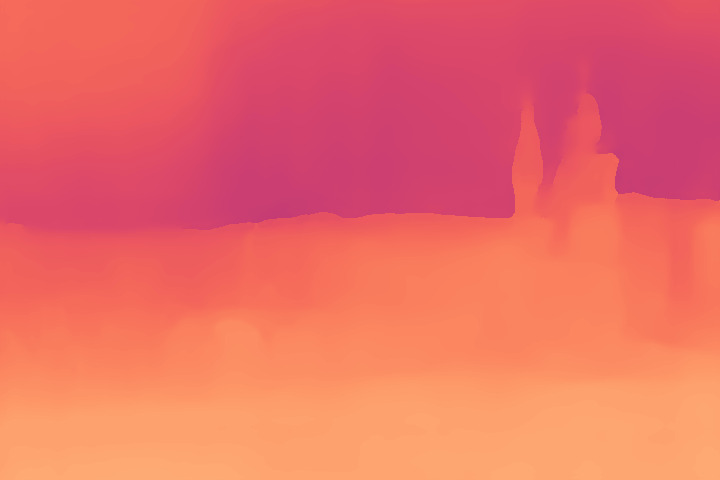} & \includegraphics[width=0.14\textwidth]{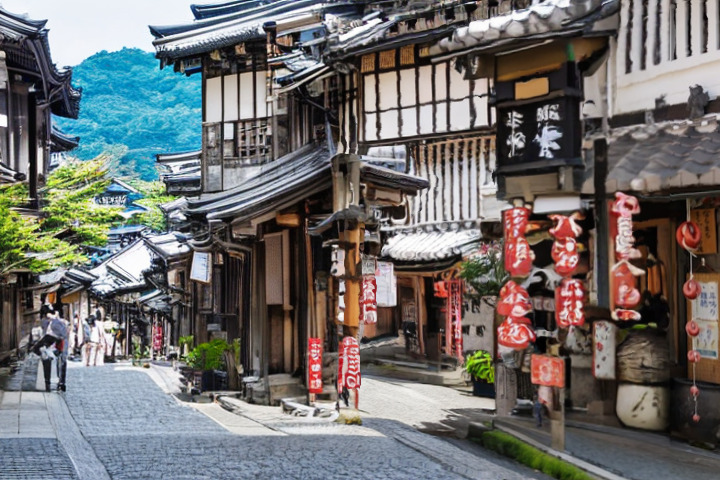} & \includegraphics[width=0.14\textwidth]{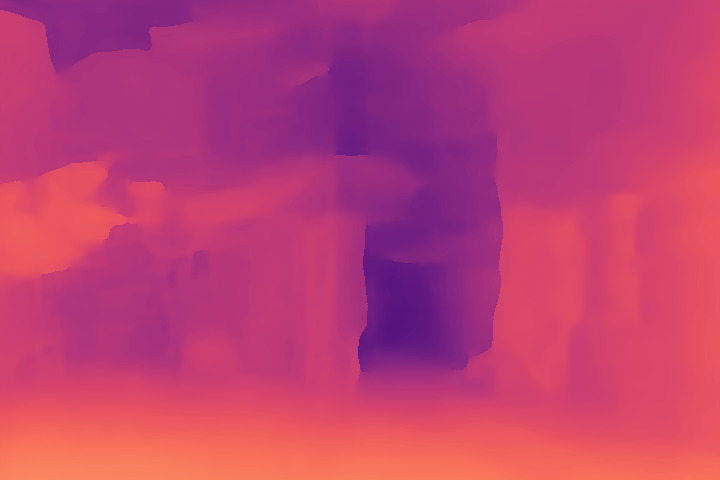} &
        \includegraphics[width=0.14\textwidth]{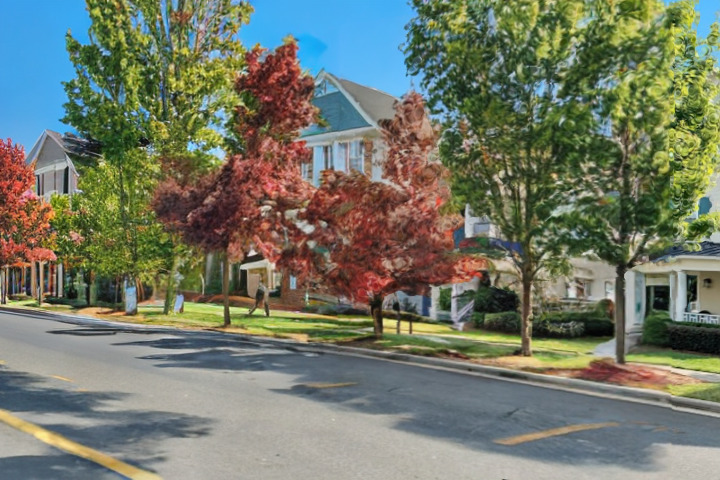} & \includegraphics[width=0.14\textwidth]{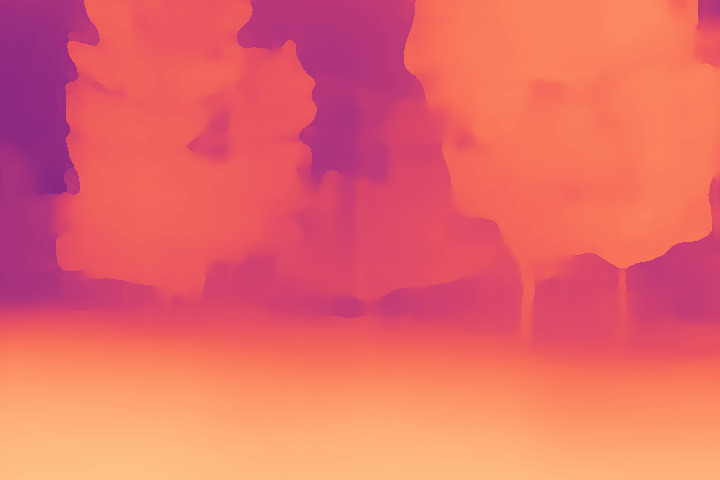} \\
        75$^{\circ}$ & \includegraphics[width=0.14\textwidth]{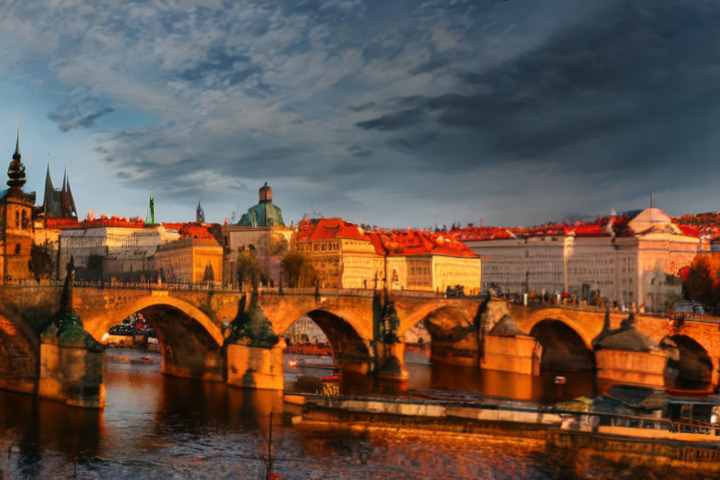} & \includegraphics[width=0.14\textwidth]{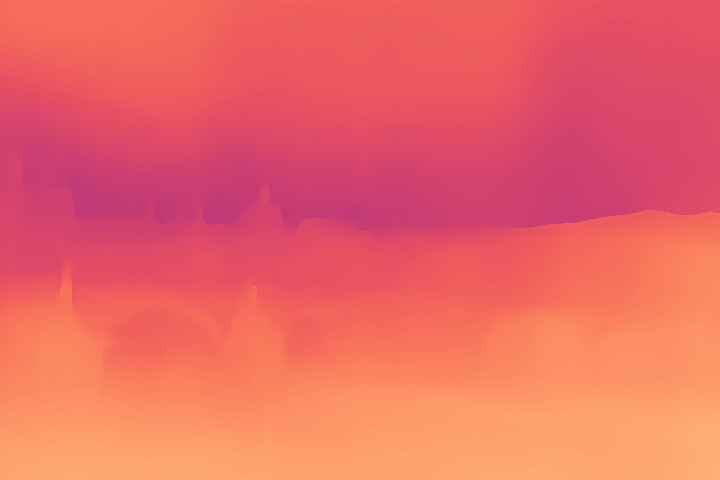} & \includegraphics[width=0.14\textwidth]{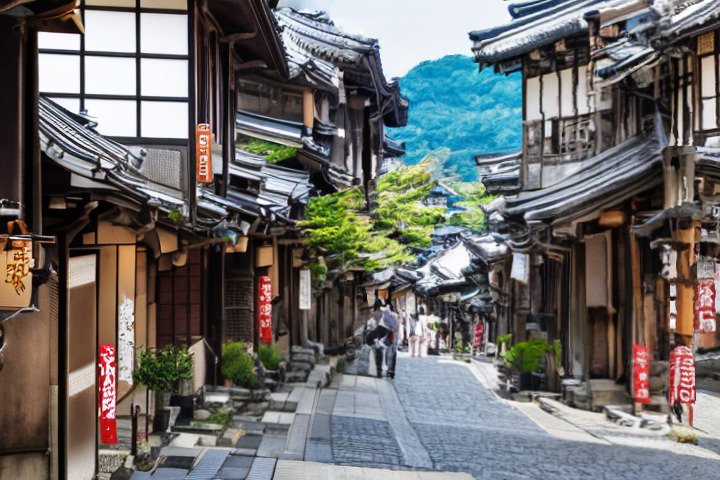} & \includegraphics[width=0.14\textwidth]{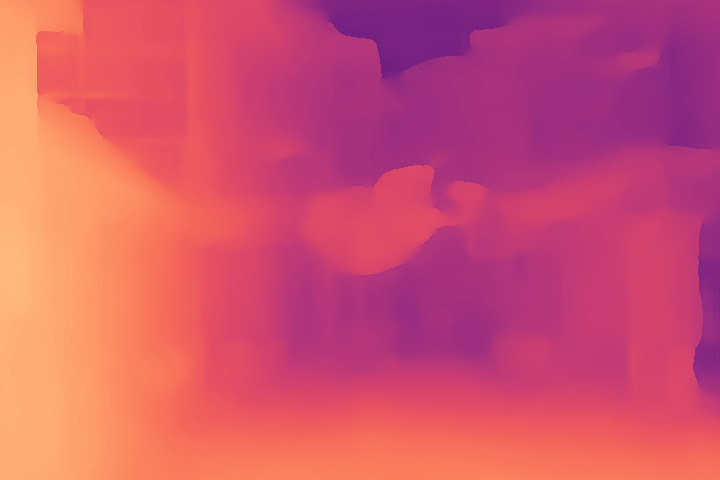} &
        \includegraphics[width=0.14\textwidth]{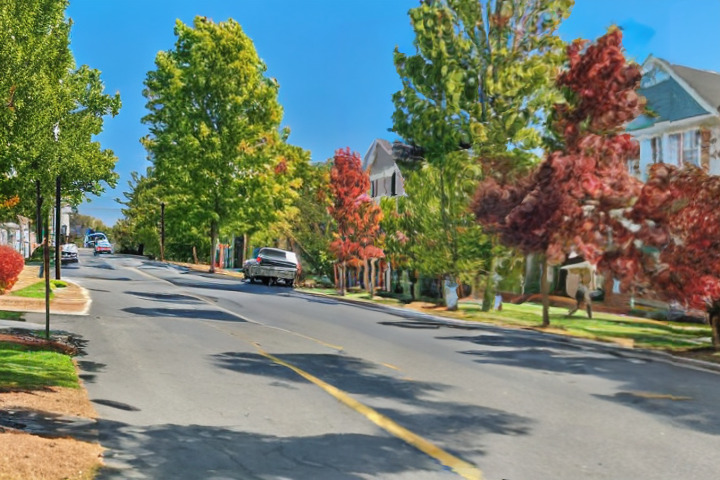} & \includegraphics[width=0.14\textwidth]{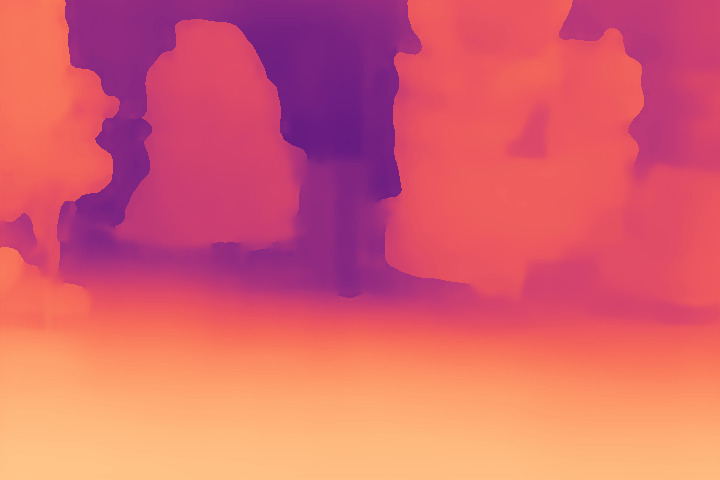} \\
        100$^{\circ}$ & \includegraphics[width=0.14\textwidth]{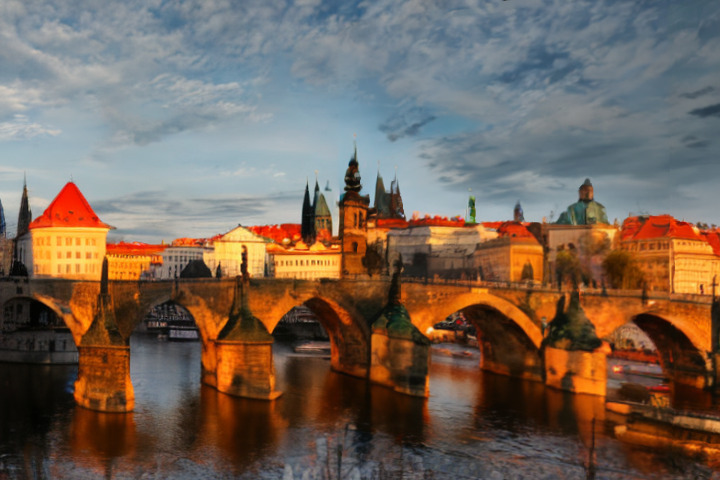} & \includegraphics[width=0.14\textwidth]{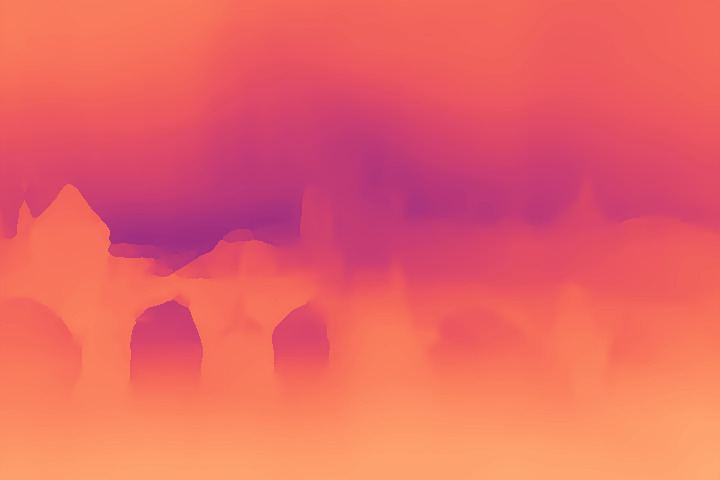} & \includegraphics[width=0.14\textwidth]{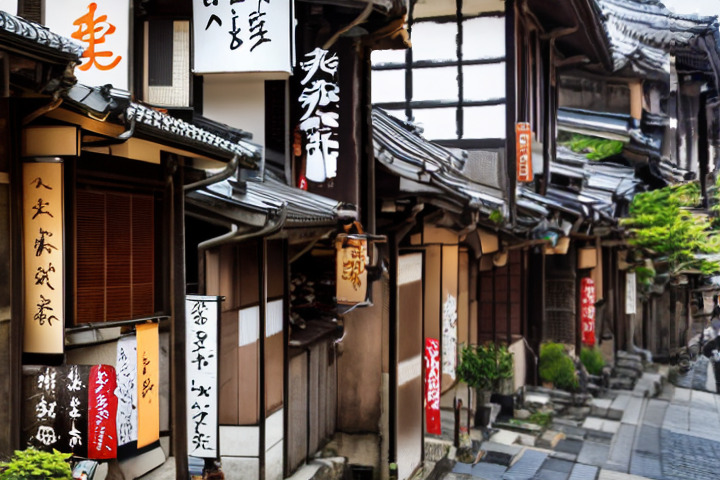} & \includegraphics[width=0.14\textwidth]{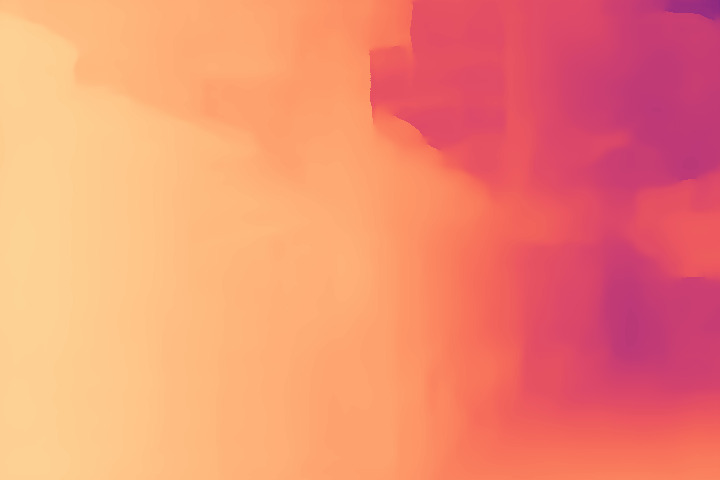} &
        \includegraphics[width=0.14\textwidth]{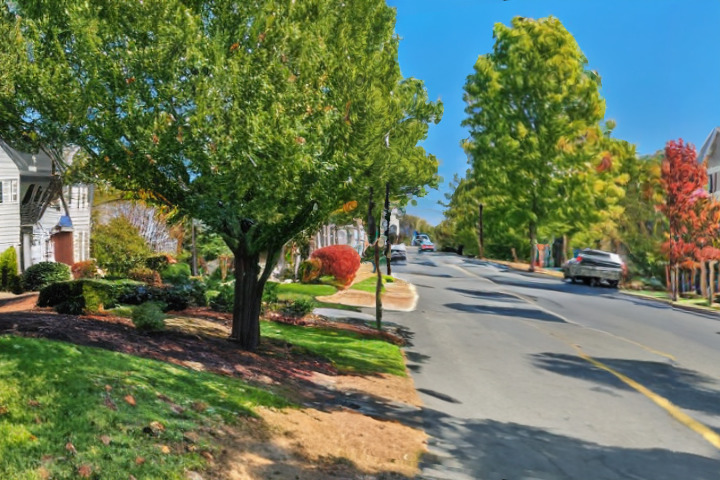} & \includegraphics[width=0.14\textwidth]{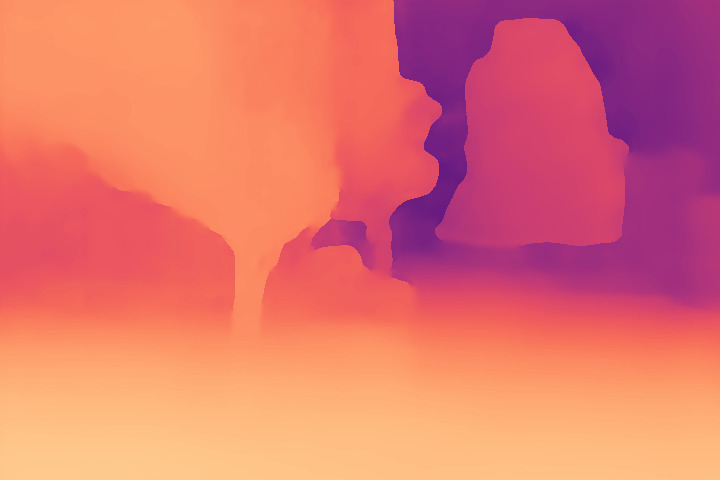} \\
        125$^{\circ}$ & \includegraphics[width=0.14\textwidth]{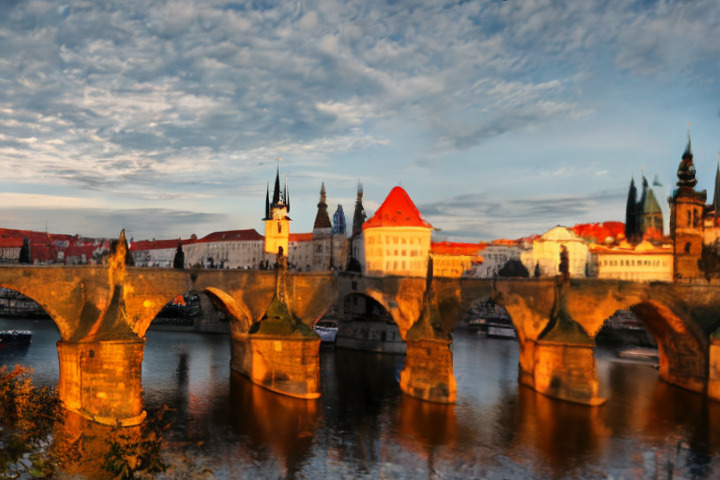} & \includegraphics[width=0.14\textwidth]{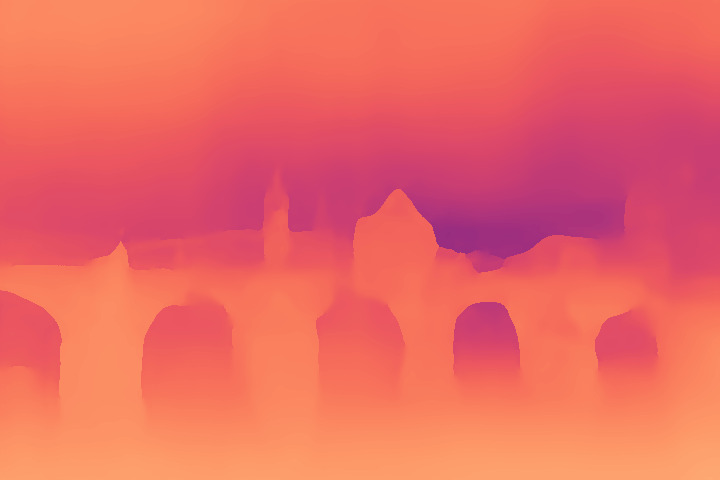} & \includegraphics[width=0.14\textwidth]{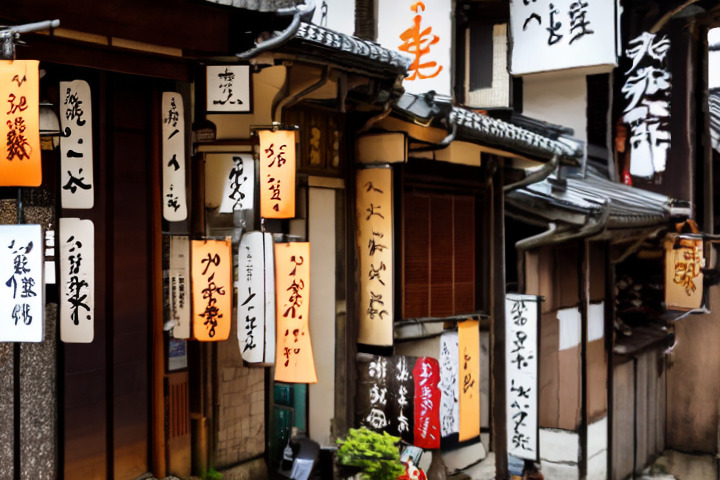} & \includegraphics[width=0.14\textwidth]{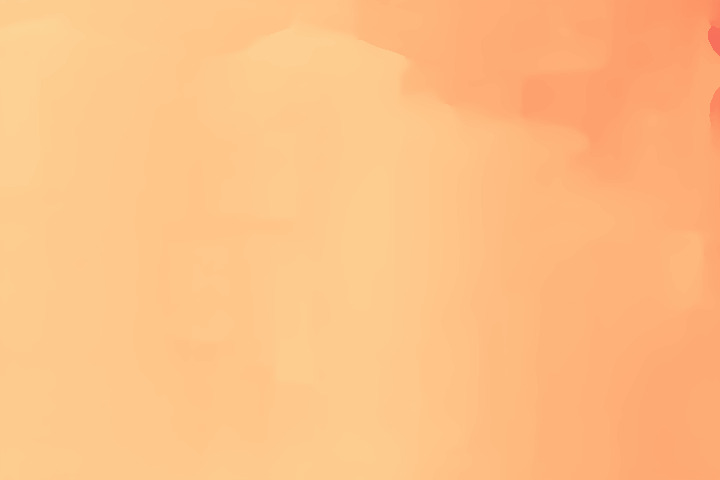} &
        \includegraphics[width=0.14\textwidth]{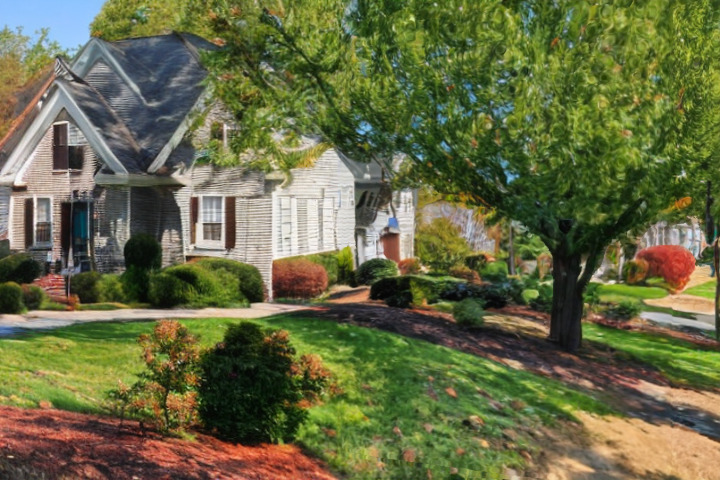} & \includegraphics[width=0.14\textwidth]{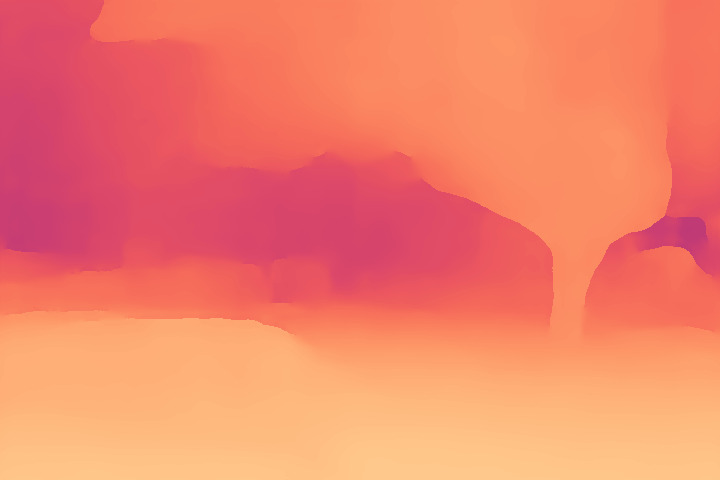} \\
        175$^{\circ}$ & \includegraphics[width=0.14\textwidth]{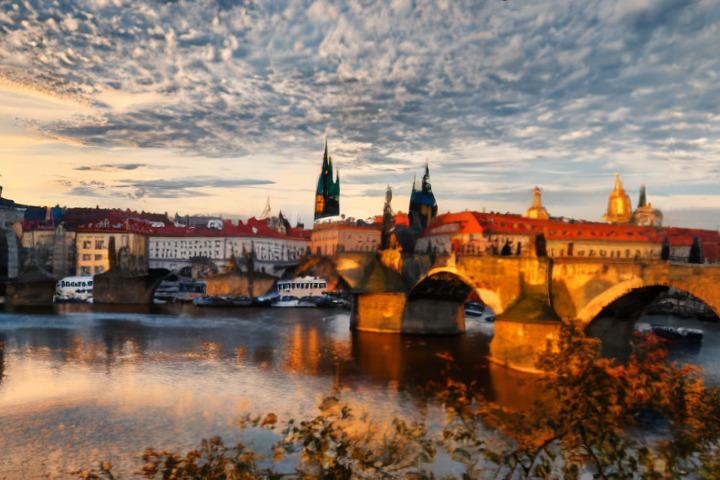} & \includegraphics[width=0.14\textwidth]{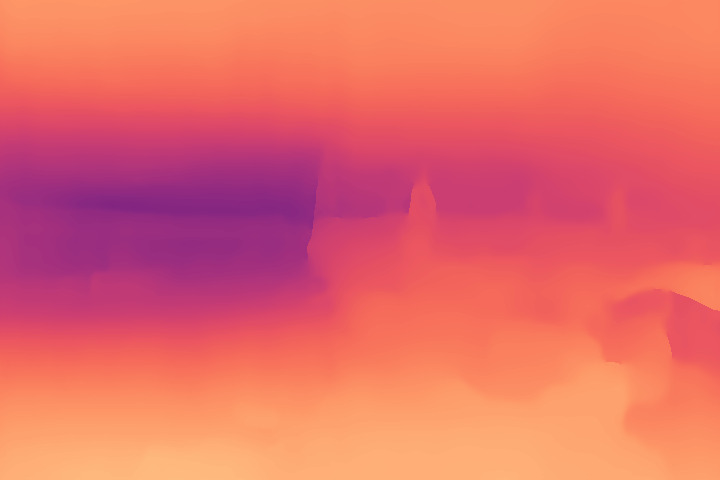} & \includegraphics[width=0.14\textwidth]{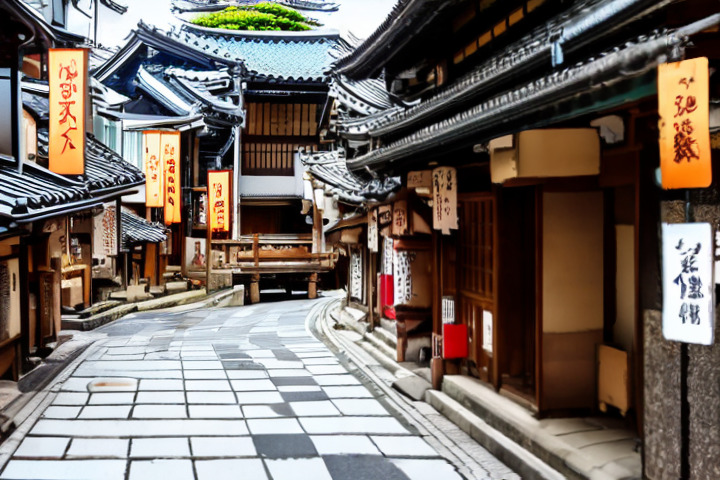} & \includegraphics[width=0.14\textwidth]{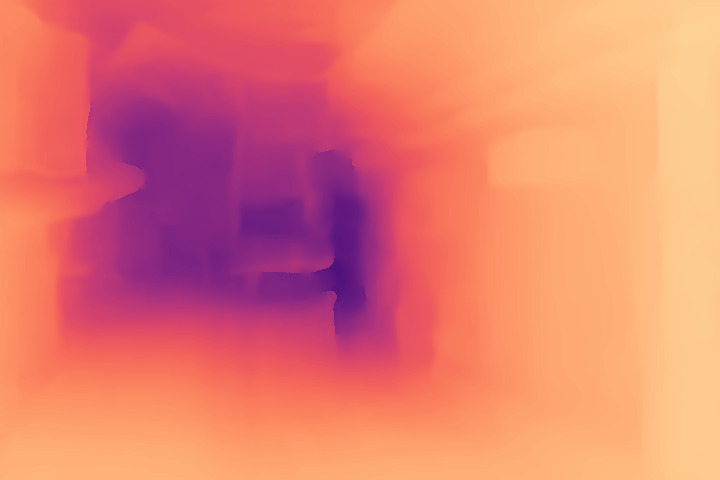} &
        \includegraphics[width=0.14\textwidth]{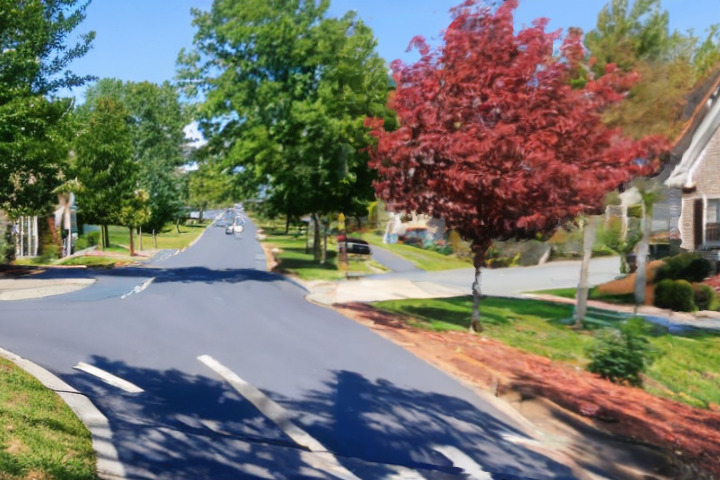} & \includegraphics[width=0.14\textwidth]{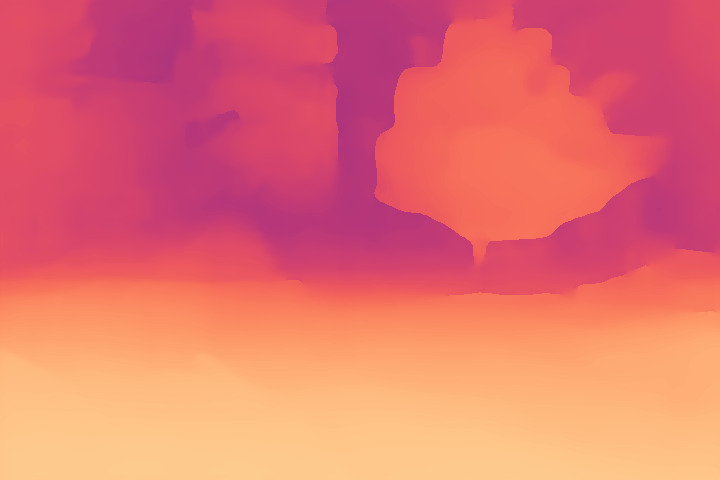} \\
        225$^{\circ}$ & \includegraphics[width=0.14\textwidth]{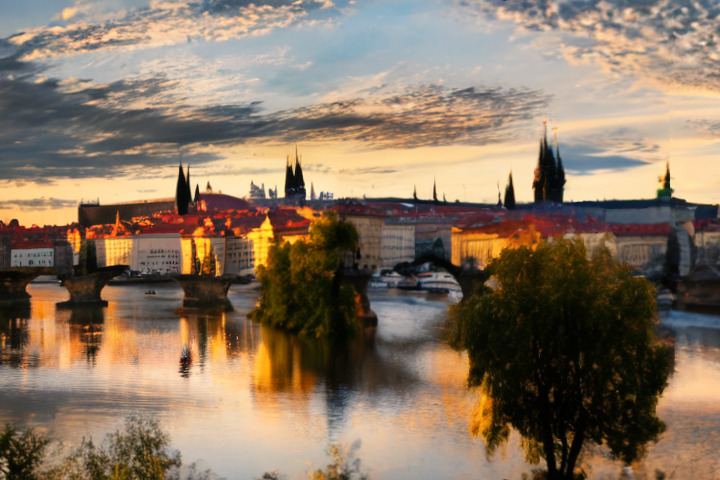} & \includegraphics[width=0.14\textwidth]{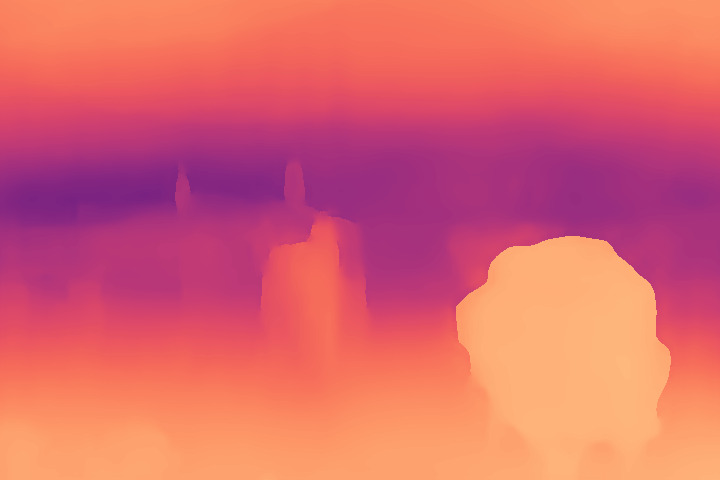} & \includegraphics[width=0.14\textwidth]{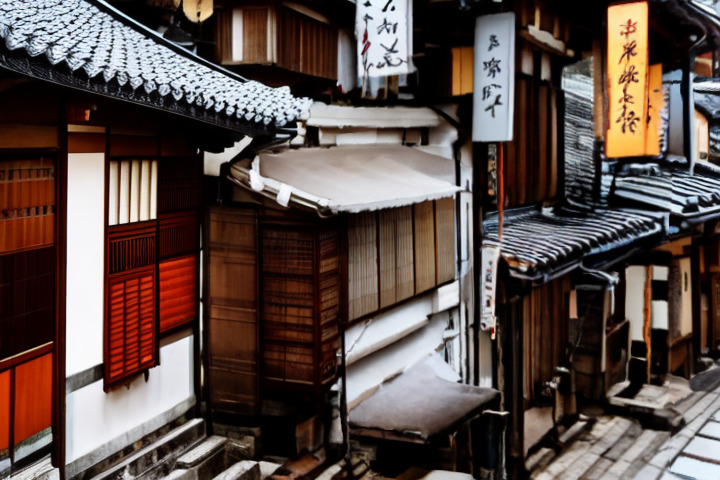} & \includegraphics[width=0.14\textwidth]{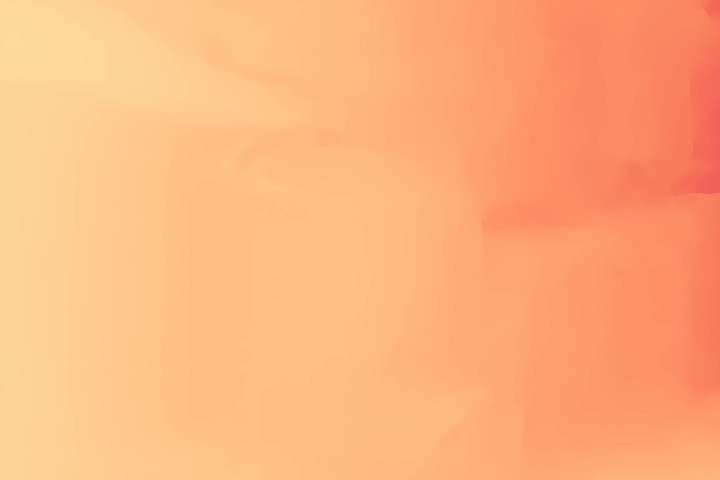} &
        \includegraphics[width=0.14\textwidth]{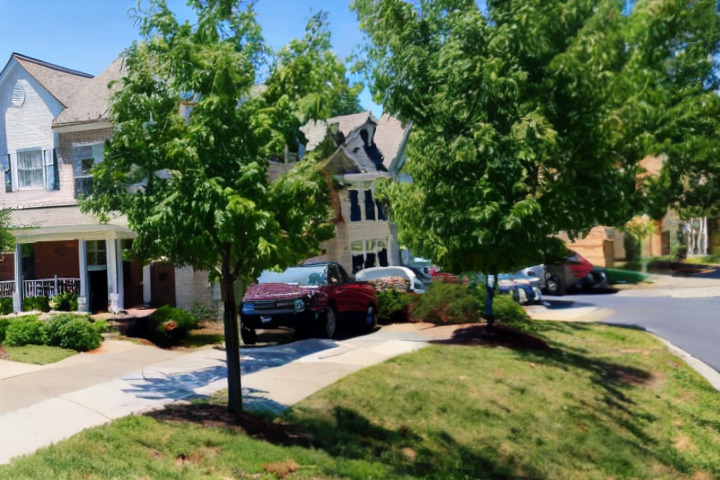} & \includegraphics[width=0.14\textwidth]{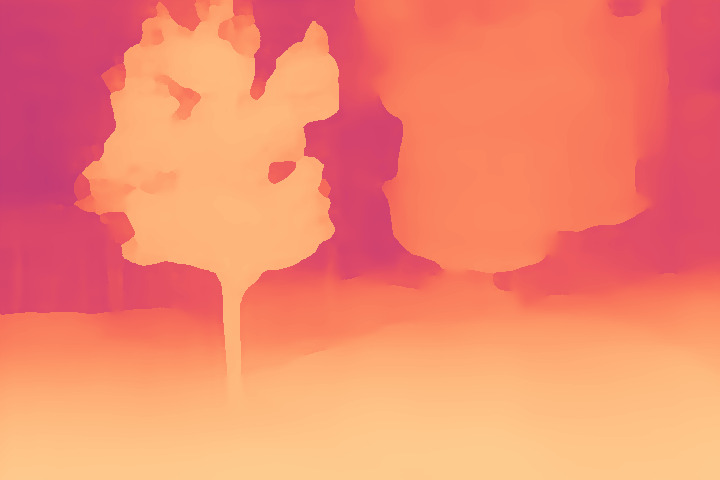} \\
        272.5$^{\circ}$ & \includegraphics[width=0.14\textwidth]{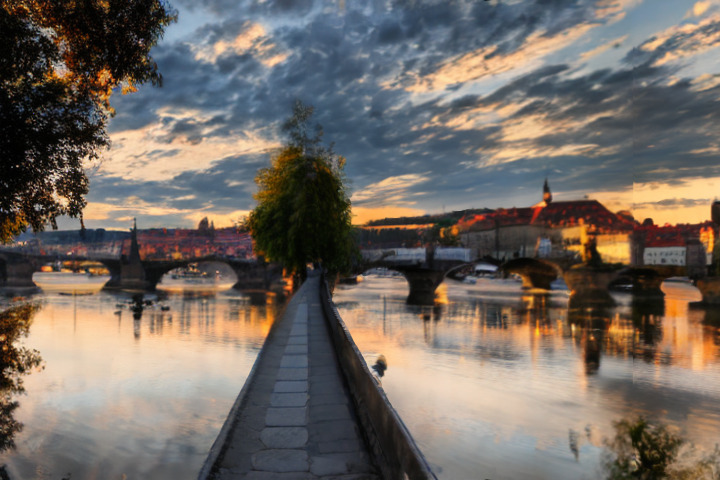} & \includegraphics[width=0.14\textwidth]{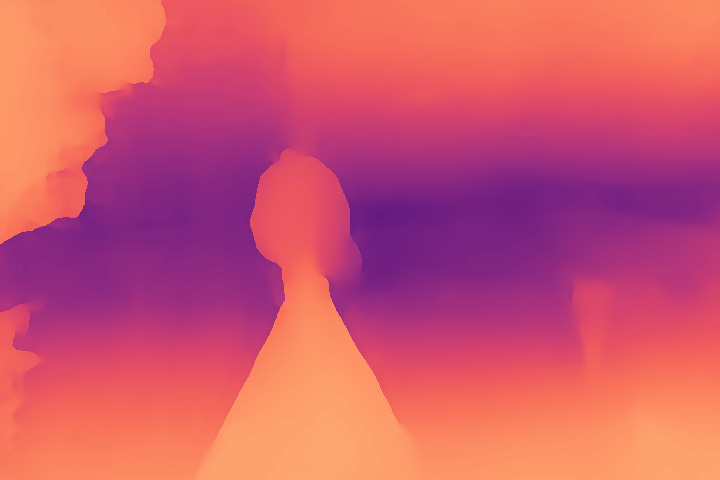} & \includegraphics[width=0.14\textwidth]{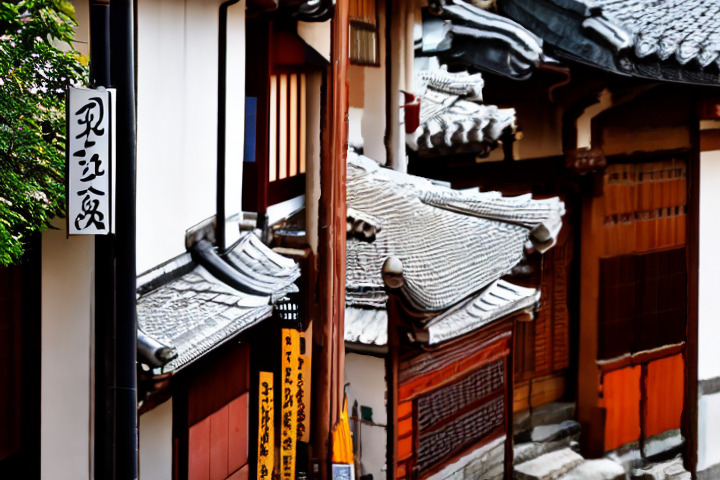} & \includegraphics[width=0.14\textwidth]{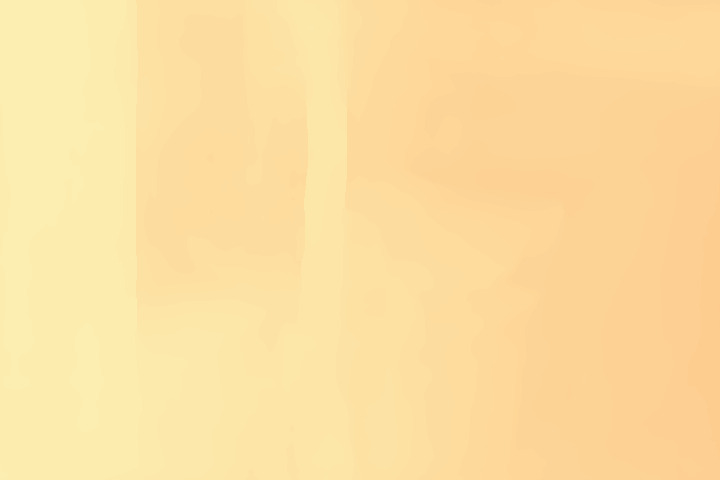} &
        \includegraphics[width=0.14\textwidth]{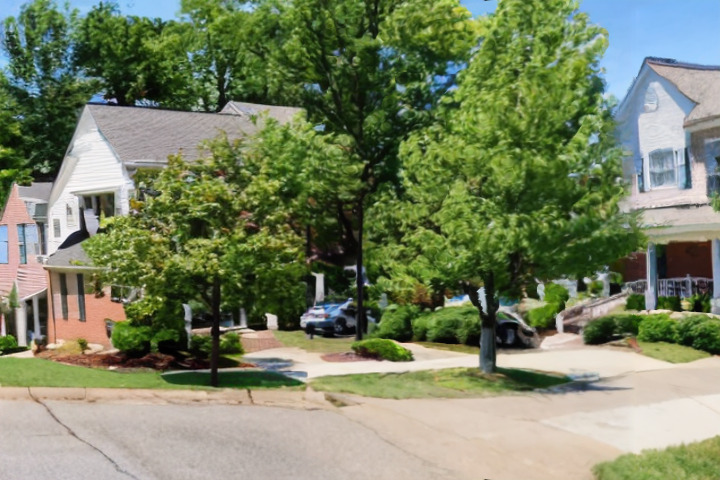} & \includegraphics[width=0.14\textwidth]{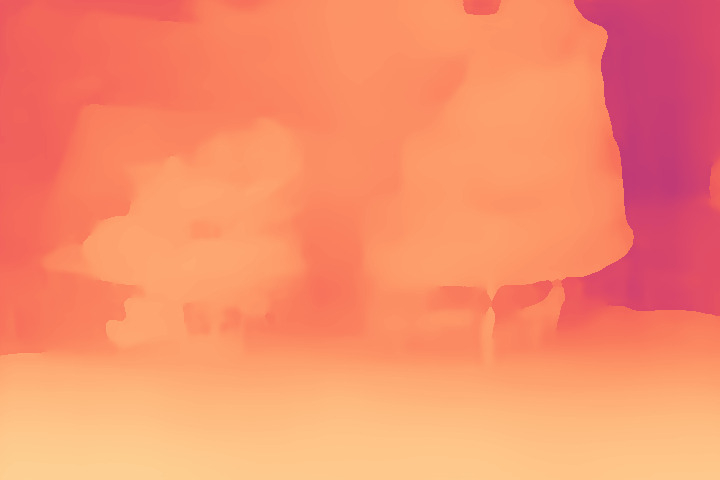} \\
        316.25$^{\circ}$ & \includegraphics[width=0.14\textwidth]{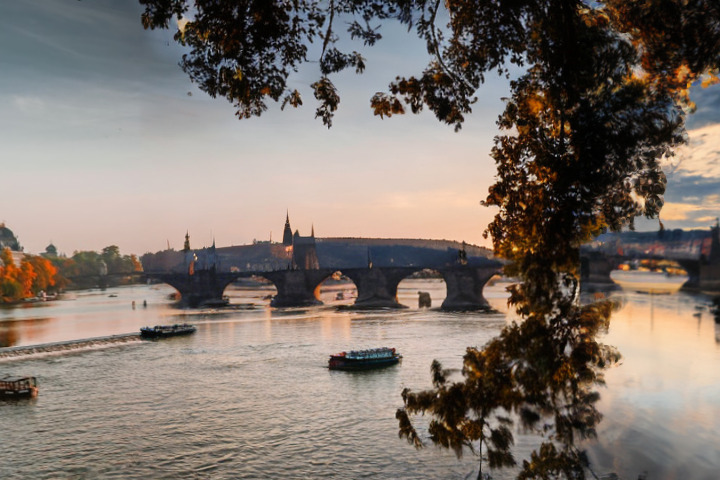} & \includegraphics[width=0.14\textwidth]{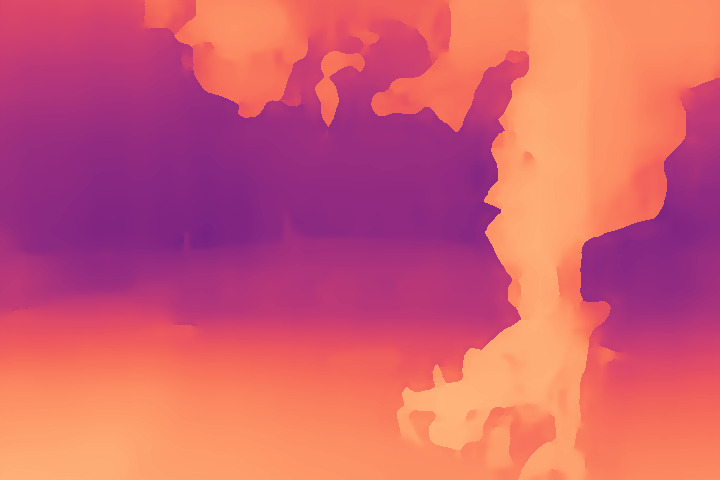} & \includegraphics[width=0.14\textwidth]{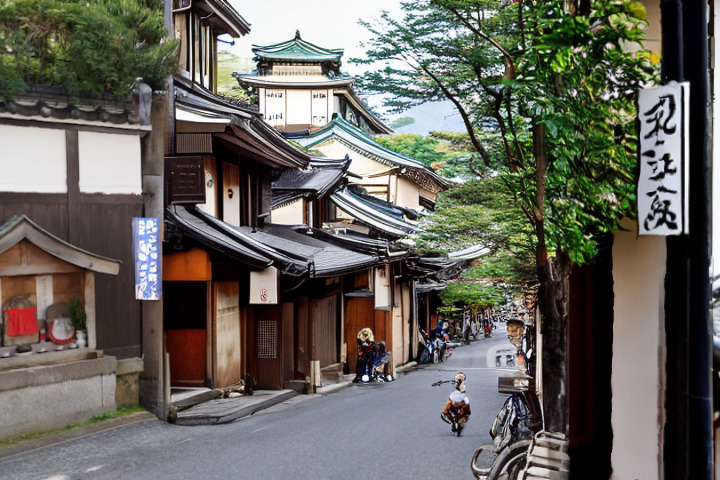} & \includegraphics[width=0.14\textwidth]{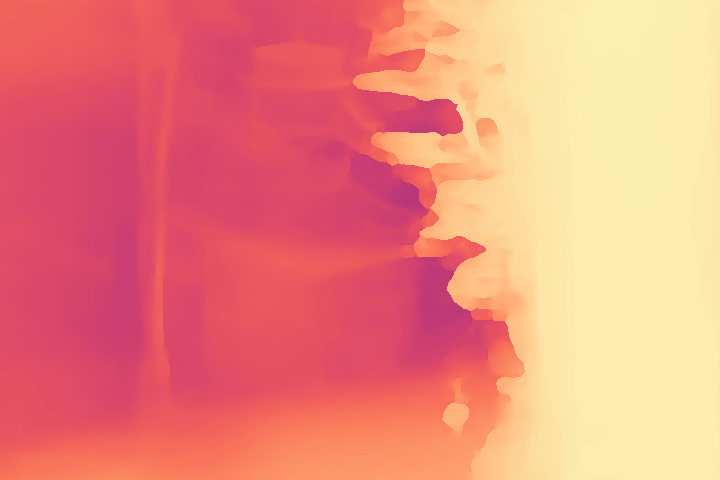} &
        \includegraphics[width=0.14\textwidth]{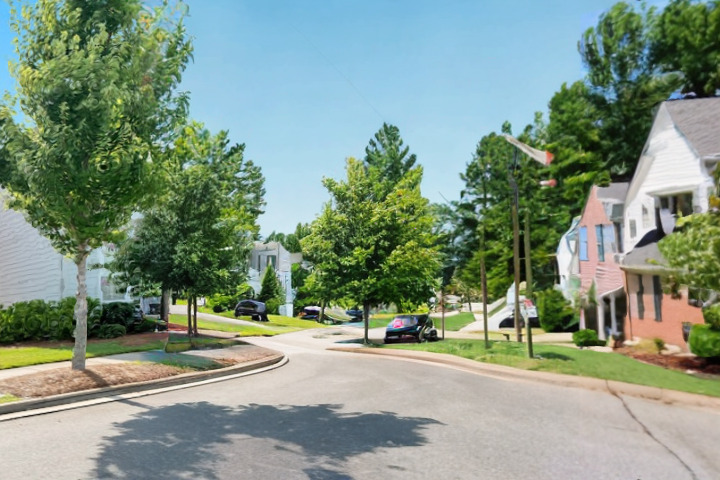} & \includegraphics[width=0.14\textwidth]{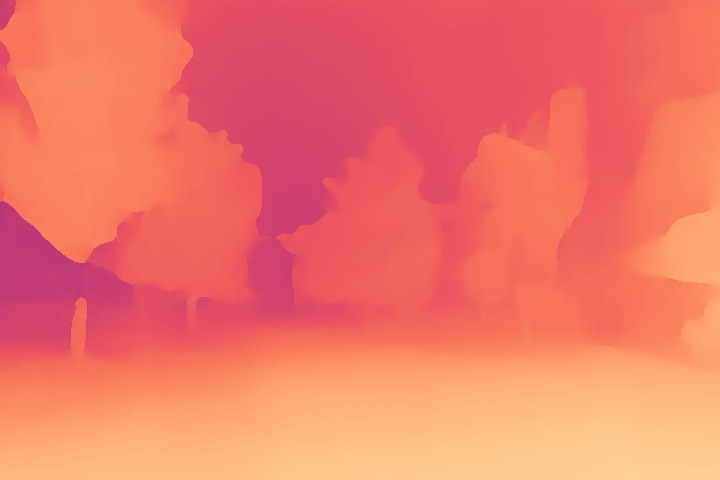}
        \\[0.5cm]
        3D Scene & \multicolumn{2}{c}{\includegraphics[width=0.28\textwidth]{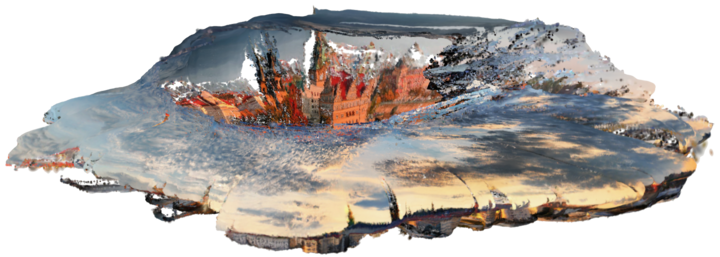}} & \multicolumn{2}{c}{\includegraphics[width=0.28\textwidth]{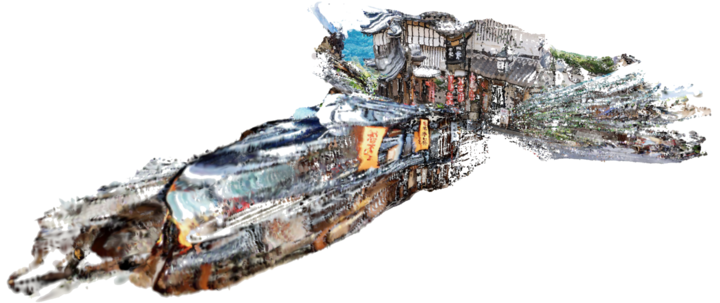}} &
        \multicolumn{2}{c}{\includegraphics[width=0.28\textwidth]{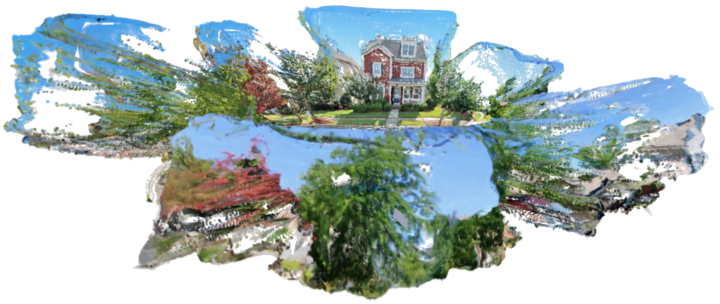}} \\
        Cut-Away & \multicolumn{2}{c}{\includegraphics[width=0.28\textwidth]{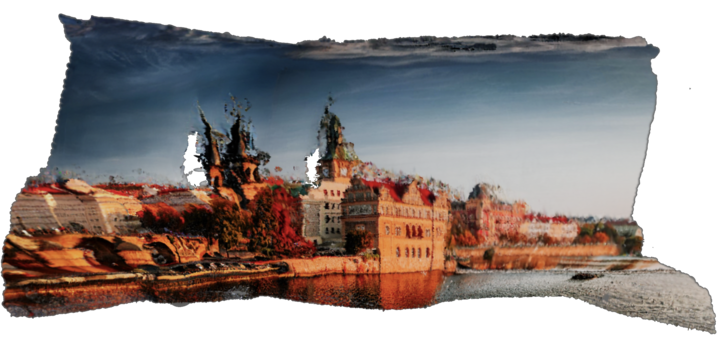}} & \multicolumn{2}{c}{\includegraphics[width=0.28\textwidth]{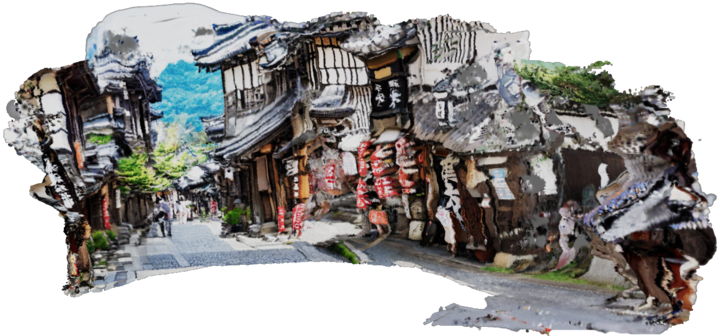}} &
        \multicolumn{2}{c}{\includegraphics[width=0.28\textwidth]{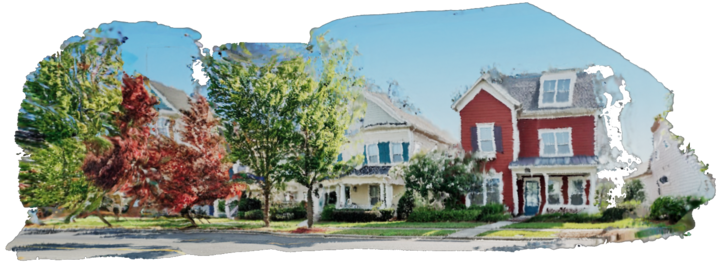}} \\
    \end{tabular}
    \caption{\textbf{Qualitative results of our method on real-world images.} We show hallucinated views and the corresponding depth maps of 360-degree scenes. We also provide a full view of the generated 360-degree scene as well as a more detailed cut-away view.}
    \label{fig:qualitative-results}
\end{figure}

Using this pipeline, we can generate 360-degree scenes given a single input image as well as a text prompt. In \Cref{fig:qualitative-results}, we show three example scenes generated by our approach that feature complex geometry. Our depth inpainting model is able to seamlessly extend scenes with believable geometry, creating an immersive experience.

We generate these results by rotating the camera 25 degrees along its azimuth with each step, slightly tapering it towards the end to close the loop. To aid the Gaussian splat optimization, we use eight support views per original view, letting the supporting view's look-at camera peek behind objects by allowing a relative change of the azimuth and elevation angles of up to $\pm 5$ degrees.

\subsubsection{Visual Quality}
\begin{table}[]
    \centering
    \begin{tabular}{l @{\hspace{10\tabcolsep}} c @{\hspace{5\tabcolsep}} c @{\hspace{5\tabcolsep}}c}
    \toprule
        \multirow{2}{*}{Method} & \multicolumn{3}{c}{CLIP-Score \cite{hessel2021clipscore}} \\
               & Prague & Kyoto & North Carolina \\
        \midrule
        LucidDreamer~\cite{chung23luciddreamer:} & 26.01 & 29.56 & 24.96 \\
        Ours & 26.87 & 27.65 & 24.24 \\
        \bottomrule
    \end{tabular}
    \caption{\textbf{Quantitative results for the visual quality of generated scenes.} Both methods receive the same input image and text prompt (see \Cref{fig:qualitative-results}) to generate 360-degree scenes. We use all hallucinated views to compute the average CLIP-Score.}
    \label{tab:clip-score}
\end{table}

To compare the visual quality of generated scenes with LucidDreamer~\cite{chung23luciddreamer:}, we utilize the CLIP-Score. We use the same input images and prompts shown in \Cref{fig:qualitative-results} to produce similar scenes with both methods. As LucidDreamer~\cite{chung23luciddreamer:} and our method both use Stable Diffusion to hallucinate images, the results we present for this metric presented in \Cref{tab:clip-score} carry limited information.

\subsection{Evaluating the Scene Geometry}
To add another dimension to our evaluation, we detail the results of our proposed scene geometry evaluation benchmark (see \Cref{sec:sgeb}) for LucidDreamer~\cite{chung23luciddreamer:}, the state-of-the-art depth completion network CostDCNet~\cite{kam2022costdcnet}, ZoeDepth~\cite{bhat2023zoedepth}, as well as our fine-tuned inpainting network in \Cref{tab:sce}.

With ZoeDepth, we predict depth and run a global scale-and-shift alignment optimization with the existing scene representation. This approach is representative of the depth estimation methods used by other current scene generation methods like Text2Immersion \cite{ouyang23text2immersion:}, Text2Room \cite{hollein23text2room:}, and WonderJourney \cite{yu23wonderjourney:}. We note that while LucidDreamer is similar to these methods, i.e., it is  uses a monocular depth estimation network and a global alignment procedure, it adds another step to smoothly connect the predicted depth with the existing scene: It interpolates the depth values between the existing scene and the predicted depth to avoid seams. We observe that this operation appears to come at the cost of geometric consistency.

In both, a real-world and a photorealistic setting, our inpainting model produces predictions that are more faithful to the ground-truth than the other methods.

\begin{table}[]
    \centering
    \begin{tabular}{l @{\hspace{10\tabcolsep}} cc}
    \toprule
        Method & ScanNet~\cite{dai2017scannet} & Hypersim~\cite{roberts2021hypersim} \\
        \midrule
        LucidDreamer~\cite{chung23luciddreamer:} & 0.1604 & 0.8057 \\
        CostDCNet~\cite{kam2022costdcnet} & 0.5854 & 4.0149 \\
        \midrule
        ZoeDepth~\cite{bhat2023zoedepth} & 0.1293 & 0.7872 \\
        \midrule
        Ours & \textbf{0.0816} & \textbf{0.7295} \\
        \bottomrule
    \end{tabular}
    \caption{\textbf{Scene geometry evaluation results of scene-generation methods on a real-world and a photorealistic dataset.} We report the average depth reconstruction error produced by our proposed depth-inpainting method as well as related approaches. The notable difference in the errors between ScanNet and Hypersim can be explained by the complexity and simulated nature of Hypersim: First, it features notably more fine details than ScanNet that cannot be recovered by models operating at a lower resolution than its image size, such as ZoeDepth. Second, unlike real-world depth sensors, the depth in Hypersim is exact with sharp boundaries, which makes it more difficult for models trained on real-world data.}
    \label{tab:sce}
\end{table}

\subsection{Ablations}

To validate the effectiveness of the design choices in our training pipeline, we ablate them and provide their results on our scene geometry evaluation benchmark in \Cref{tab:ablations-table}.

\begin{table}[]
    \centering
    \begin{tabular}{ccccccc}
    \toprule
        Input & Depth Annot. & $p$ & Warped Masks & Align. & ScanNet~\cite{dai2017scannet} & Hypersim~\cite{roberts2021hypersim} \\
        \midrule
        RGB+sd & Original & 0.5 & - & - & 0.7734 & 2.2913 \\
        \midrule
        RGB+sd & Original & 0.5 & \checkmark & - & 0.1015 & 0.7615 \\
        RGB+sd & ZoeDepth & 0.5 & \checkmark & - & 0.0793 & 0.7555 \\
        \midrule
        RGB+sd & Marigold & 0.0 & \checkmark & - & 0.0864 & 0.7547 \\
        RGB+sd & Marigold & 0.25 & \checkmark & - & \textbf{0.0791} & 0.7301 \\
        RGB+sd & Marigold & 0.75 & \checkmark & - & 0.0869 & 0.7578 \\
        \midrule
        RGB & Marigold & 0.5 & \checkmark & - & 0.2553 & 1.1536 \\
        RGB & Marigold & 0.5 & \checkmark & \checkmark & 0.1335 & 0.8152 \\
        \midrule
        RGB+sd & Marigold & 0.5 & \checkmark & - & 0.0816 & \textbf{0.7295} \\
        \bottomrule
    \end{tabular}
    \caption{\textbf{Scene geometry evaluation results for ablations of our method.} We consider the input for our model (image-only or supplemented with sparse depth), the source of depth annotations in our fine-tuning process to learn the inpainting task, the probability $p$ that we mask out the sparse depth input, whether we use warped masks during the fine-tuning process that mimic characteristic inpainting patterns in scene generation, and if the final depth prediction is aligned with the existing point cloud through a global scale-and-shift operation.}
    \label{tab:ablations-table}
\end{table}

\subsubsection{Distillation of High-Resolution Models}
It generally seems beneficial for the fine-tuning process to obtain a stronger training signal for depth annotations that are fully dense. Due to its simulated nature, Hypersim has ground-truth depth maps with notably finer structures than ScanNet, which was captured with less precise real-world tools. With Marigold~\cite{ke2023repurposing} being trained on Hypersim, it is able to reproduce these high-resolution details. As we observe improved performance for this dataset once we use Marigold predictions as ground-truth, we deem our knowledge distillation setup to be effective.

\subsubsection{Inpainting Task Probability}
We observe that there is merit to not allocating too little or too much time in the training procedure to learning the inpainting task. Not dedicating any time to the original depth estimation task without any sparse input appears to negatively impact the performance. We find that spending between 50-75\% (i.e., $p \in [0.25, 0.5]$) of the time in the fine-tuning process training the inpainting task yields performant models.

\subsubsection{Masking Strategy}
We observe that using warped masks that mimic the characteristic inpainting patterns that occur when changing view points is critical to yield a high-performing depth inpainting model for the task of depth inpainting in a scene generation setting. A naive patch-based masking approach, where regions of varying size are randomly masked, produces inferior results.

\subsubsection{Zero Sparse Depth Input}
In our fine-tuning setup, we can set a probability $p$ to zero out the sparse depth input to the model, effectively reverting to the monocular depth estimation task. We observe that in this setting, our model is highly competitive with the original ZoeDepth model (see \Cref{tab:sce}), suggesting the inpainting ability has been bolted onto the network in our fine-tuning setup with only minor degradation of the original task. However, this setting is not able to reach the performance of a network that is given sparse depth input. This supports our hypothesis that adding sparse depth information of the existing scene leads to predictions that are geometrically more coherent and overall more faithful to observed depth.

\section{Conclusion}
We introduce a scene geometry evaluation benchmark that allows assessing the geometric quality of scene generation methods. While generated scenes are difficult to evaluate due to the lack of ground-truth data, we provide a rigorous method to uncover geometric inconsistencies. We also propose a general-purpose depth inpainting model that completes depth maps, highly suitable for the inpainting challenges arising in a scene generation task. Compared to previous methods, this network produces depth predictions that are more consistent with existing scenes. Finally, we showcase the network as part of a 360-degree scene generation pipeline, where it seamlessly stitches newly hallucinated frames to existing ones, allowing it to generate compelling and immersive 3D scenes.

\paragraph{Ethics.}
For further details on ethics, data protection, and copyright please see \url{https://www.robots.ox.ac.uk/~vedaldi/research/union/ethics.html}.

\paragraph{Acknowledgements.}
P. E., A. V., I. L., and C.R. are supported by ERC-UNION- CoG-101001212. P.E. is also supported by Meta Research. I.L. and C.R. also receive support from VisualAI EP/T028572/1.

\clearpage
\bibliographystyle{splncs04}
\bibliography{main}

\begin{thebibliography}{10}
\providecommand{\url}[1]{\texttt{#1}}
\providecommand{\urlprefix}{URL }
\providecommand{\doi}[1]{https://doi.org/#1}

\bibitem{bautista22gaudi:}
Bautista, M.{\'{A}}., Guo, P., Abnar, S., Talbott, W., Toshev, A., Chen, Z., Dinh, L., Zhai, S., Goh, H., Ulbricht, D., Dehghan, A., Susskind, J.M.: {GAUDI:} {A} neural architect for immersive 3d scene generation  \textbf{abs/2207.13751} (2022)

\bibitem{bhat2023zoedepth}
Bhat, S.F., Birkl, R., Wofk, D., Wonka, P., M{\"u}ller, M.: Zoedepth: Zero-shot transfer by combining relative and metric depth. arXiv preprint arXiv:2302.12288  (2023)

\bibitem{boulch2020fkaconv}
Boulch, A., Puy, G., Marlet, R.: Fkaconv: Feature-kernel alignment for point cloud convolution. In: Proceedings of the Asian Conference on Computer Vision (2020)

\bibitem{cai23diffdreamer:}
Cai, S., Chan, E.R., Peng, S., Shahbazi, M., Obukhov, A., Gool, L.V., Wetzstein, G.: {DiffDreamer}: Towards consistent unsupervised single-view scene extrapolation with conditional diffusion models. In: ICCV (2023)

\bibitem{chan2022efficient}
Chan, E.R., Lin, C.Z., Chan, M.A., Nagano, K., Pan, B., De~Mello, S., Gallo, O., Guibas, L.J., Tremblay, J., Khamis, S., et~al.: Efficient geometry-aware 3d generative adversarial networks. In: Proceedings of the IEEE/CVF Conference on Computer Vision and Pattern Recognition. pp. 16123--16133 (2022)

\bibitem{chan2021pi}
Chan, E.R., Monteiro, M., Kellnhofer, P., Wu, J., Wetzstein, G.: pi-gan: Periodic implicit generative adversarial networks for 3d-aware image synthesis. In: Proceedings of the IEEE/CVF conference on computer vision and pattern recognition. pp. 5799--5809 (2021)

\bibitem{chen2019learning}
Chen, Y., Yang, B., Liang, M., Urtasun, R.: Learning joint 2d-3d representations for depth completion. In: Proceedings of the IEEE/CVF International Conference on Computer Vision. pp. 10023--10032 (2019)

\bibitem{cheng2020cspn}
Cheng, X., Wang, P., Guan, C., Yang, R.: Cspn++: Learning context and resource aware convolutional spatial propagation networks for depth completion. In: Proceedings of the AAAI Conference on Artificial Intelligence. vol.~34, pp. 10615--10622 (2020)

\bibitem{cheng2018depth}
Cheng, X., Wang, P., Yang, R.: Depth estimation via affinity learned with convolutional spatial propagation network. In: Proceedings of the European conference on computer vision (ECCV). pp. 103--119 (2018)

\bibitem{chung23luciddreamer:}
Chung, J., Lee, S., Nam, H., Lee, J., Lee, K.M.: Luciddreamer: Domain-free generation of 3d gaussian splatting scenes  \textbf{abs/2311.13384} (2023)

\bibitem{cohen-bar23set-the-scene:}
Cohen{-}Bar, D., Richardson, E., Metzer, G., Giryes, R., Cohen{-}Or, D.: Set-the-scene: Global-local training for generating controllable nerf scenes (2023)

\bibitem{dai2017scannet}
Dai, A., Chang, A.X., Savva, M., Halber, M., Funkhouser, T., Nie{\ss}ner, M.: Scannet: Richly-annotated 3d reconstructions of indoor scenes. In: Proceedings of the IEEE conference on computer vision and pattern recognition. pp. 5828--5839 (2017)

\bibitem{deng2022depth}
Deng, K., Liu, A., Zhu, J.Y., Ramanan, D.: Depth-supervised nerf: Fewer views and faster training for free. In: Proceedings of the IEEE/CVF Conference on Computer Vision and Pattern Recognition. pp. 12882--12891 (2022)

\bibitem{eigen2014depth}
Eigen, D., Puhrsch, C., Fergus, R.: Depth map prediction from a single image using a multi-scale deep network. Advances in neural information processing systems  \textbf{27} (2014)

\bibitem{gu2021stylenerf}
Gu, J., Liu, L., Wang, P., Theobalt, C.: Stylenerf: A style-based 3d-aware generator for high-resolution image synthesis. arXiv preprint arXiv:2110.08985  (2021)

\bibitem{hessel2021clipscore}
Hessel, J., Holtzman, A., Forbes, M., Bras, R.L., Choi, Y.: {CLIPScore:} a reference-free evaluation metric for image captioning. In: EMNLP (2021)

\bibitem{hollein23text2room:}
H{\"{o}}llein, L., Cao, A., Owens, A., Johnson, J., Nie{\ss}ner, M.: {Text2Room}: Extracting textured {3D} meshes from {2D} text-to-image models. In: ICCV (2023)

\bibitem{hu21worldsheet:}
Hu, R., Ravi, N., Berg, A.C., Pathak, D.: Worldsheet: Wrapping the world in a 3d sheet for view synthesis from a single image. In: ICCV (2021)

\bibitem{huynh2021boosting}
Huynh, L., Nguyen, P., Matas, J., Rahtu, E., Heikkil{\"a}, J.: Boosting monocular depth estimation with lightweight 3d point fusion. In: Proceedings of the IEEE/CVF International Conference on Computer Vision. pp. 12767--12776 (2021)

\bibitem{jain2021putting}
Jain, A., Tancik, M., Abbeel, P.: Putting nerf on a diet: Semantically consistent few-shot view synthesis. In: Proceedings of the IEEE/CVF International Conference on Computer Vision. pp. 5885--5894 (2021)

\bibitem{kam2022costdcnet}
Kam, J., Kim, J., Kim, S., Park, J., Lee, S.: Costdcnet: Cost volume based depth completion for a single rgb-d image. In: European Conference on Computer Vision. pp. 257--274. Springer (2022)

\bibitem{ke2023repurposing}
Ke, B., Obukhov, A., Huang, S., Metzger, N., Daudt, R.C., Schindler, K.: Repurposing diffusion-based image generators for monocular depth estimation. arXiv preprint arXiv:2312.02145  (2023)

\bibitem{kerbl20233d}
Kerbl, B., Kopanas, G., Leimk{\"u}hler, T., Drettakis, G.: 3d gaussian splatting for real-time radiance field rendering. ACM Transactions on Graphics  \textbf{42}(4) (2023)

\bibitem{lei23rgbd2:}
Lei, J., Tang, J., Jia, K.: {RGBD2:} generative scene synthesis via incremental view inpainting using {RGBD} diffusion models. In: CVPR (2023)

\bibitem{li21mine:}
Li, J., Feng, Z., She, Q., Ding, H., Wang, C., Lee, G.H.: {MINE:} towards continuous depth {MPI} with nerf for novel view synthesis. In: ICCV (2021)

\bibitem{li2022infinitenature}
Li, Z., Wang, Q., Snavely, N., Kanazawa, A.: Infinitenature-zero: Learning perpetual view generation of natural scenes from single images. In: European Conference on Computer Vision. pp. 515--534. Springer (2022)

\bibitem{lin2023magic3d}
Lin, C.H., Gao, J., Tang, L., Takikawa, T., Zeng, X., Huang, X., Kreis, K., Fidler, S., Liu, M.Y., Lin, T.Y.: Magic3d: High-resolution text-to-3d content creation. In: Proceedings of the IEEE/CVF Conference on Computer Vision and Pattern Recognition. pp. 300--309 (2023)

\bibitem{liu2021infinite}
Liu, A., Tucker, R., Jampani, V., Makadia, A., Snavely, N., Kanazawa, A.: Infinite nature: Perpetual view generation of natural scenes from a single image. In: Proceedings of the IEEE/CVF International Conference on Computer Vision. pp. 14458--14467 (2021)

\bibitem{liu2023zero}
Liu, R., Wu, R., Van~Hoorick, B., Tokmakov, P., Zakharov, S., Vondrick, C.: Zero-1-to-3: Zero-shot one image to 3d object. In: Proceedings of the IEEE/CVF International Conference on Computer Vision. pp. 9298--9309 (2023)

\bibitem{melas2023realfusion}
Melas-Kyriazi, L., Laina, I., Rupprecht, C., Vedaldi, A.: Realfusion: 360deg reconstruction of any object from a single image. In: Proceedings of the IEEE/CVF Conference on Computer Vision and Pattern Recognition. pp. 8446--8455 (2023)

\bibitem{mildenhall19local}
Mildenhall, B., Srinivasan, P.P., Cayon, R.O., Kalantari, N.K., Ramamoorthi, R., Ng, R., Kar, A.: Local light field fusion: practical view synthesis with prescriptive sampling guidelines  \textbf{38}(4) (2019)

\bibitem{mildenhall2021nerf}
Mildenhall, B., Srinivasan, P.P., Tancik, M., Barron, J.T., Ramamoorthi, R., Ng, R.: Nerf: Representing scenes as neural radiance fields for view synthesis. Communications of the ACM  \textbf{65}(1),  99--106 (2021)

\bibitem{Silberman:ECCV12}
Nathan~Silberman, Derek~Hoiem, P.K., Fergus, R.: Indoor segmentation and support inference from rgbd images. In: ECCV (2012)

\bibitem{nguyen2019hologan}
Nguyen-Phuoc, T., Li, C., Theis, L., Richardt, C., Yang, Y.L.: Hologan: Unsupervised learning of 3d representations from natural images. In: Proceedings of the IEEE/CVF International Conference on Computer Vision. pp. 7588--7597 (2019)

\bibitem{niemeyer2021giraffe}
Niemeyer, M., Geiger, A.: Giraffe: Representing scenes as compositional generative neural feature fields. In: Proceedings of the IEEE/CVF Conference on Computer Vision and Pattern Recognition. pp. 11453--11464 (2021)

\bibitem{ouyang23text2immersion:}
Ouyang, H., Heal, K., Lombardi, S., Sun, T.: Text2immersion: Generative immersive scene with 3d gaussians  \textbf{abs/2312.09242} (2023)

\bibitem{park2020non}
Park, J., Joo, K., Hu, Z., Liu, C.K., So~Kweon, I.: Non-local spatial propagation network for depth completion. In: Computer Vision--ECCV 2020: 16th European Conference, Glasgow, UK, August 23--28, 2020, Proceedings, Part XIII 16. pp. 120--136. Springer (2020)

\bibitem{poole2022dreamfusion}
Poole, B., Jain, A., Barron, J.T., Mildenhall, B.: Dreamfusion: Text-to-3d using 2d diffusion. arXiv preprint arXiv:2209.14988  (2022)

\bibitem{qian2023magic123}
Qian, G., Mai, J., Hamdi, A., Ren, J., Siarohin, A., Li, B., Lee, H.Y., Skorokhodov, I., Wonka, P., Tulyakov, S., et~al.: Magic123: One image to high-quality 3d object generation using both 2d and 3d diffusion priors. arXiv preprint arXiv:2306.17843  (2023)

\bibitem{raj2023dreambooth3d}
Raj, A., Kaza, S., Poole, B., Niemeyer, M., Ruiz, N., Mildenhall, B., Zada, S., Aberman, K., Rubinstein, M., Barron, J., et~al.: Dreambooth3d: Subject-driven text-to-3d generation. arXiv preprint arXiv:2303.13508  (2023)

\bibitem{roberts2021hypersim}
Roberts, M., Ramapuram, J., Ranjan, A., Kumar, A., Bautista, M.A., Paczan, N., Webb, R., Susskind, J.M.: Hypersim: A photorealistic synthetic dataset for holistic indoor scene understanding. In: Proceedings of the IEEE/CVF international conference on computer vision. pp. 10912--10922 (2021)

\bibitem{rockwell21pixelsynth:}
Rockwell, C., Fouhey, D.F., Johnson, J.: {PixelSynth}: Generating a {3D}-consistent experience from a single image. In: ICCV (2021)

\bibitem{rombach2022high}
Rombach, R., Blattmann, A., Lorenz, D., Esser, P., Ommer, B.: High-resolution image synthesis with latent diffusion models. In: Proceedings of the IEEE/CVF conference on computer vision and pattern recognition. pp. 10684--10695 (2022)

\bibitem{sargent2023zeronvs}
Sargent, K., Li, Z., Shah, T., Herrmann, C., Yu, H.X., Zhang, Y., Chan, E.R., Lagun, D., Fei-Fei, L., Sun, D., et~al.: Zeronvs: Zero-shot 360-degree view synthesis from a single real image. arXiv preprint arXiv:2310.17994  (2023)

\bibitem{saxena2023surprising}
Saxena, S., Herrmann, C., Hur, J., Kar, A., Norouzi, M., Sun, D., Fleet, D.J.: The surprising effectiveness of diffusion models for optical flow and monocular depth estimation. arXiv preprint arXiv:2306.01923  (2023)

\bibitem{seitz97photorealistic}
Seitz, S.M., Dyer, C.R.: Photorealistic scene reconstruction by voxel coloring. In: CVPR (1997)

\bibitem{shi2023mvdream}
Shi, Y., Wang, P., Ye, J., Long, M., Li, K., Yang, X.: Mvdream: Multi-view diffusion for 3d generation. arXiv preprint arXiv:2308.16512  (2023)

\bibitem{shih203d-photography}
Shih, M., Su, S., Kopf, J., Huang, J.: 3d photography using context-aware layered depth inpainting. In: CVPR (2020)

\bibitem{sitzmann19deepvoxels:}
Sitzmann, V., Thies, J., Heide, F., Nie{\ss}ner, M., Wetzstein, G., Zollh{\"{o}}fer, M.: Deepvoxels: Learning persistent 3d feature embeddings. In: CVPR (2019)

\bibitem{stan2023ldm3d}
Stan, G.B.M., Wofk, D., Fox, S., Redden, A., Saxton, W., Yu, J., Aflalo, E., Tseng, S.Y., Nonato, F., Muller, M., et~al.: Ldm3d: Latent diffusion model for 3d. arXiv preprint arXiv:2305.10853  (2023)

\bibitem{szymanowicz2023splatter}
Szymanowicz, S., Rupprecht, C., Vedaldi, A.: Splatter image: Ultra-fast single-view 3d reconstruction. arXiv preprint arXiv:2312.13150  (2023)

\bibitem{trevithick21grf:}
Trevithick, A., Yang, B.: {GRF:} learning a general radiance field for 3d representation and rendering. In: ICCV (2021)

\bibitem{tulsiani18layer-structured}
Tulsiani, S., Tucker, R., Snavely, N.: Layer-structured 3d scene inference via view synthesis. In: ECCV (2018)

\bibitem{wang2018deep}
Wang, S., Suo, S., Ma, W.C., Pokrovsky, A., Urtasun, R.: Deep parametric continuous convolutional neural networks. In: Proceedings of the IEEE conference on computer vision and pattern recognition. pp. 2589--2597 (2018)

\bibitem{wang2024prolificdreamer}
Wang, Z., Lu, C., Wang, Y., Bao, F., Li, C., Su, H., Zhu, J.: Prolificdreamer: High-fidelity and diverse text-to-3d generation with variational score distillation. Advances in Neural Information Processing Systems  \textbf{36} (2024)

\bibitem{wiles20synsin:}
Wiles, O., Gkioxari, G., Szeliski, R., Johnson, J.: Synsin: End-to-end view synthesis from a single image. In: CVPR (2020)

\bibitem{wu24blockfusion:}
Wu, Z., Li, Y., Yan, H., Shang, T., Sun, W., Wang, S., Cui, R., Liu, W., Sato, H., Li, H., Ji, P.: {BlockFusion}: Expandable {3D} scene generation using latent tri-plane extrapolation  (2024)

\bibitem{xiang233d-aware}
Xiang, J., Yang, J., Huang, B., Tong, X.: {3D}-aware image generation using {2D} diffusion models  \textbf{abs/2303.17905} (2023)

\bibitem{yu2021pixelnerf}
Yu, A., Ye, V., Tancik, M., Kanazawa, A.: pixelnerf: Neural radiance fields from one or few images. In: Proceedings of the IEEE/CVF Conference on Computer Vision and Pattern Recognition. pp. 4578--4587 (2021)

\bibitem{yu23wonderjourney:}
Yu, H., Duan, H., Hur, J., Sargent, K., Rubinstein, M., Freeman, W.T., Cole, F., Sun, D., Snavely, N., Wu, J., Herrmann, C.: Wonderjourney: Going from anywhere to everywhere  \textbf{abs/2312.03884} (2023)

\bibitem{zhang23text2nerf:}
Zhang, J., Li, X., Wan, Z., Wang, C., Liao, J.: Text2nerf: Text-driven 3d scene generation with neural radiance fields  \textbf{abs/2305.11588} (2023)

\bibitem{zhang2023completionformer}
Zhang, Y., Guo, X., Poggi, M., Zhu, Z., Huang, G., Mattoccia, S.: Completionformer: Depth completion with convolutions and vision transformers. In: Proceedings of the IEEE/CVF Conference on Computer Vision and Pattern Recognition. pp. 18527--18536 (2023)

\bibitem{zhou2017places}
Zhou, B., Lapedriza, A., Khosla, A., Oliva, A., Torralba, A.: Places: A 10 million image database for scene recognition. IEEE transactions on pattern analysis and machine intelligence  \textbf{40}(6),  1452--1464 (2017)

\bibitem{zhu2023designing}
Zhu, Z., Feng, X., Chen, D., Bao, J., Wang, L., Chen, Y., Yuan, L., Hua, G.: Designing a better asymmetric vqgan for stablediffusion. arXiv preprint arXiv:2306.04632  (2023)

\end{thebibliography}
\end{document}


\maketitle

\section{Overview}
This supplementary document is organized as follows: First, we provide more detailed implementation details with our specific choices of hyperparameters for the fine-tuning process of our model and the scene generation pipeline (\cref{sec:details}). Second, we provide further qualitative results produced by our scene generation pipeline, including a new set of scenes for different real-world images (\cref{sec:qualitative}).
Third, we show that our depth completion model can be plugged into an existing scene generation method to improve the structural quality of the scenes it generates (\cref{sec:improving}).
Finally, we discuss the limitations of our as well as related approaches in 3D scene generation (\cref{sec:limitations}). We hope that this discussion will inspire future research to further this field.

We invite the reader to consider the accompanying videos that show renderings of 3D scenes generated with our method. Please note that the videos have a higher resolution (1080p) than the frames used to generated the scenes (720p).

We will publish the code to train our depth completion model, the trained checkpoint, as well as our pipeline to generate 3D scenes.

\section{Further Implementation Details}
\label{sec:details}
\subsection{Fine-Tuning}
We base our model on ZoeDepth, which uses a dense prediction transformer (DPT)~\cite{ranftl2021vision} with a BeiT (Bidirectional Encoder representation from Image Transformers)~\cite{bao2021beit} backbone at a resolution of $512 \times 384$. We fine-tune the model for 5 epochs with batch size 8, using a low learning rate of 0.00025 with a weight decay of 0.01. We train on four NVIDIA Tesla P40 GPUs.

\subsection{Scene Generation}
For the first two phases of our scene generation pipeline described in Section 5.2, we use PyTorch3D~\cite{ravi2020accelerating} to render our point cloud representation. We use alpha-composite rendering with 16 points per camera ray at an image size of $720 \times 480$. The point cloud generated by the first two phases provides the starting point for the Gaussian splat optimization~\cite{kerbl20233d}, alongside images of the generated and supporting views. We use the default set of parameters for the optimization but limit the number of iterations to 2,990.

\subsection{Ablations}
In Section 5.4, we present ablations of the design choices in our training pipeline. To demonstrate the benefit of providing a sparse depth input to our model, we compare it to a scenario where the model is only given an image as input, reverting to the original depth estimation task. While the sparse depth inherently provides information about the overall depth of the scene, this information is missing if only an image is given as input. To ensure a fair comparison, we thus run a global scale-and-shift optimization on the predicted depth to align it with the existing scene depth. This optimization is run for at most 100 steps, stopping early if no improvement is made for 10 steps. It is based on the absolute error between both depth maps (where available) and uses the Adam optimizer with a learning rate of 0.01.

\section{Additional Qualitative Results} 
\label{sec:qualitative}
We present an additional set of generated scenes from real-world images in \Cref{fig:supp-qualitative-results}. As in Figure 4, we provide individual images of the hallucinated views, a rendering of the entire generated 3D scene, as well as a cut-away that provides more detail.

In \Cref{fig:supp-diversity}, we provide further examples where we generate a scene multiple times from the same input image, using different seeds for Stable Diffusion.

\begin{figure}
    \centering
    \begin{tabular}{c @{\hspace{5\tabcolsep}} cc @{\hspace{5\tabcolsep}} cc @{\hspace{5\tabcolsep}} cc}
     Prompt & \multicolumn{2}{c}{\makecell{Beautiful buildings in \\ a quaint UK village}} & \multicolumn{2}{c}{\makecell{View over Cairo, Egypt}} & \multicolumn{2}{c}{\makecell{Mountains in Peru \\ on an overcast day}} \\[0.5cm]
        Input & \includegraphics[width=0.14\textwidth]{images/qualitative_results/westminster_street/p1_outpainted_view_0.png} & \includegraphics[width=0.14\textwidth]{images/qualitative_results/westminster_street/p1_outpainted_depth_0.png} & \includegraphics[width=0.14\textwidth]{images/qualitative_results/cairo/p1_outpainted_view_0.png} & \includegraphics[width=0.14\textwidth]{images/qualitative_results/cairo/p1_outpainted_depth_0.png} & \includegraphics[width=0.14\textwidth]{images/qualitative_results/machu_picchu/p1_outpainted_view_0.png} & \includegraphics[width=0.14\textwidth]{images/qualitative_results/machu_picchu/p1_outpainted_depth_0.png} \\
        25$^{\circ}$ & \includegraphics[width=0.14\textwidth]{images/qualitative_results/westminster_street/p1_outpainted_view_25.png} & \includegraphics[width=0.14\textwidth]{images/qualitative_results/westminster_street/p1_outpainted_depth_25.png} & \includegraphics[width=0.14\textwidth]{images/qualitative_results/cairo/p1_outpainted_view_25.png} & \includegraphics[width=0.14\textwidth]{images/qualitative_results/cairo/p1_outpainted_depth_25.png} &
        \includegraphics[width=0.14\textwidth]{images/qualitative_results/machu_picchu/p1_outpainted_view_25.png} & \includegraphics[width=0.14\textwidth]{images/qualitative_results/machu_picchu/p1_outpainted_depth_25.png} \\
        50$^{\circ}$ & \includegraphics[width=0.14\textwidth]{images/qualitative_results/westminster_street/p1_outpainted_view_50.png} & \includegraphics[width=0.14\textwidth]{images/qualitative_results/westminster_street/p1_outpainted_depth_50.png} & \includegraphics[width=0.14\textwidth]{images/qualitative_results/cairo/p1_outpainted_view_50.png} & \includegraphics[width=0.14\textwidth]{images/qualitative_results/cairo/p1_outpainted_depth_50.png} &
        \includegraphics[width=0.14\textwidth]{images/qualitative_results/machu_picchu/p1_outpainted_view_50.png} & \includegraphics[width=0.14\textwidth]{images/qualitative_results/machu_picchu/p1_outpainted_depth_50.png} \\
        75$^{\circ}$ & \includegraphics[width=0.14\textwidth]{images/qualitative_results/westminster_street/p1_outpainted_view_75.png} & \includegraphics[width=0.14\textwidth]{images/qualitative_results/westminster_street/p1_outpainted_depth_75.png} & \includegraphics[width=0.14\textwidth]{images/qualitative_results/cairo/p1_outpainted_view_75.png} & \includegraphics[width=0.14\textwidth]{images/qualitative_results/cairo/p1_outpainted_depth_75.png} &
        \includegraphics[width=0.14\textwidth]{images/qualitative_results/machu_picchu/p1_outpainted_view_75.png} & \includegraphics[width=0.14\textwidth]{images/qualitative_results/machu_picchu/p1_outpainted_depth_75.png} \\
        100$^{\circ}$ & \includegraphics[width=0.14\textwidth]{images/qualitative_results/westminster_street/p1_outpainted_view_100.png} & \includegraphics[width=0.14\textwidth]{images/qualitative_results/westminster_street/p1_outpainted_depth_100.png} & \includegraphics[width=0.14\textwidth]{images/qualitative_results/cairo/p1_outpainted_view_100.png} & \includegraphics[width=0.14\textwidth]{images/qualitative_results/cairo/p1_outpainted_depth_100.png} &
        \includegraphics[width=0.14\textwidth]{images/qualitative_results/machu_picchu/p1_outpainted_view_100.png} & \includegraphics[width=0.14\textwidth]{images/qualitative_results/machu_picchu/p1_outpainted_depth_100.png} \\
        125$^{\circ}$ & \includegraphics[width=0.14\textwidth]{images/qualitative_results/westminster_street/p1_outpainted_view_125.png} & \includegraphics[width=0.14\textwidth]{images/qualitative_results/westminster_street/p1_outpainted_depth_125.png} & \includegraphics[width=0.14\textwidth]{images/qualitative_results/cairo/p1_outpainted_view_125.png} & \includegraphics[width=0.14\textwidth]{images/qualitative_results/cairo/p1_outpainted_depth_125.png} &
        \includegraphics[width=0.14\textwidth]{images/qualitative_results/machu_picchu/p1_outpainted_view_125.png} & \includegraphics[width=0.14\textwidth]{images/qualitative_results/machu_picchu/p1_outpainted_depth_125.png} \\
        175$^{\circ}$ & \includegraphics[width=0.14\textwidth]{images/qualitative_results/westminster_street/p1_outpainted_view_175.png} & \includegraphics[width=0.14\textwidth]{images/qualitative_results/westminster_street/p1_outpainted_depth_175.png} & \includegraphics[width=0.14\textwidth]{images/qualitative_results/cairo/p1_outpainted_view_175.png} & \includegraphics[width=0.14\textwidth]{images/qualitative_results/cairo/p1_outpainted_depth_175.png} &
        \includegraphics[width=0.14\textwidth]{images/qualitative_results/machu_picchu/p1_outpainted_view_175.png} & \includegraphics[width=0.14\textwidth]{images/qualitative_results/machu_picchu/p1_outpainted_depth_175.png} \\
        225$^{\circ}$ & \includegraphics[width=0.14\textwidth]{images/qualitative_results/westminster_street/p1_outpainted_view_225.png} & \includegraphics[width=0.14\textwidth]{images/qualitative_results/westminster_street/p1_outpainted_depth_225.png} & \includegraphics[width=0.14\textwidth]{images/qualitative_results/cairo/p1_outpainted_view_225.png} & \includegraphics[width=0.14\textwidth]{images/qualitative_results/cairo/p1_outpainted_depth_225.png} &
        \includegraphics[width=0.14\textwidth]{images/qualitative_results/machu_picchu/p1_outpainted_view_225.png} & \includegraphics[width=0.14\textwidth]{images/qualitative_results/machu_picchu/p1_outpainted_depth_225.png} \\
        272.5$^{\circ}$ & \includegraphics[width=0.14\textwidth]{images/qualitative_results/westminster_street/p1_outpainted_view_272.5.png} & \includegraphics[width=0.14\textwidth]{images/qualitative_results/westminster_street/p1_outpainted_depth_272.5.png} & \includegraphics[width=0.14\textwidth]{images/qualitative_results/cairo/p1_outpainted_view_272.5.png} & \includegraphics[width=0.14\textwidth]{images/qualitative_results/cairo/p1_outpainted_depth_272.5.png} &
        \includegraphics[width=0.14\textwidth]{images/qualitative_results/machu_picchu/p1_outpainted_view_272.5.png} & \includegraphics[width=0.14\textwidth]{images/qualitative_results/machu_picchu/p1_outpainted_depth_272.5.png} \\
        316.25$^{\circ}$ & \includegraphics[width=0.14\textwidth]{images/qualitative_results/westminster_street/p1_outpainted_view_316.25.png} & \includegraphics[width=0.14\textwidth]{images/qualitative_results/westminster_street/p1_outpainted_depth_316.25.png} & \includegraphics[width=0.14\textwidth]{images/qualitative_results/cairo/p1_outpainted_view_316.25.png} & \includegraphics[width=0.14\textwidth]{images/qualitative_results/cairo/p1_outpainted_depth_316.25.png} &
        \includegraphics[width=0.14\textwidth]{images/qualitative_results/machu_picchu/p1_outpainted_view_316.25.png} & \includegraphics[width=0.14\textwidth]{images/qualitative_results/machu_picchu/p1_outpainted_depth_316.25.png}
        \\[0.5cm]
        3D Scene & \multicolumn{2}{c}{\includegraphics[width=0.28\textwidth]{images/qualitative_results/westminster_street/westminster_street_donut.png}} & \multicolumn{2}{c}{\includegraphics[width=0.28\textwidth]{images/qualitative_results/cairo/cairo_donut.png}} &
        \multicolumn{2}{c}{\includegraphics[width=0.28\textwidth]{images/qualitative_results/machu_picchu/machu_picchu_donut.png}} \\
        Cut-Away & \multicolumn{2}{c}{\includegraphics[width=0.28\textwidth]{images/qualitative_results/westminster_street/westminster_street_gs.png}} & \multicolumn{2}{c}{\includegraphics[width=0.28\textwidth]{images/qualitative_results/cairo/cairo_gs.png}} &
        \multicolumn{2}{c}{\includegraphics[width=0.28\textwidth]{images/qualitative_results/machu_picchu/machu_picchu_gs.png}} \\
    \end{tabular}
    \caption{\textbf{Qualitative results of our method on additional real-world images.} It is able to generate convincing, immersive scenes for a wide range of input images and associated prompts.}
    \label{fig:supp-qualitative-results}
\end{figure}

\begin{figure}
    \centering
    \begin{tabular}{c}
        \includegraphics[width=\textwidth]{images/diversity/kyoto_diverse_1.jpg} \\
        \includegraphics[width=\textwidth]{images/diversity/kyoto_diverse_2.jpg} \\
        \includegraphics[width=\textwidth]{images/diversity/kyoto_diverse_3.jpg} \\
    \end{tabular}
    \caption{\textbf{Diversity of generated scenes.} All scenes have been generated by our method, starting from the same input image and text prompt (consider the Kyoto scene in Figure 4). For visualization purposes, only two additional frames have been hallucinated.}
    \label{fig:supp-diversity}
\end{figure}

\section{Improving an Existing Method with Depth Completion}
\label{sec:improving}
We demonstrate that our depth completion model is able to improve the geometric consistency of existing scene generation methods by plugging it into one such method, namely LucidDreamer~\cite{chung23luciddreamer:}. We use the same input image, prompt, and fixed seed, while swapping the depth prediction backbone. The original approach uses the depth estimation network ZoeDepth, which does not utilize a sparse depth input. We consider the scenes of Figure 4, showing the results for \textit{Prague} (\Cref{fig:ld-comparison-prague}), \textit{Kyoto} (\Cref{fig:ld-comparison-kyoto}), and \textit{North Carolina} (\Cref{fig:ld-comparison-nc}). Having plugged our model into their method, we see that its generated scenes have an overall better structural quality than those produced by the original method.

\begin{figure}
    \centering
    \begin{tabular}{c}
        \includegraphics[width=\textwidth]{images/ld_plugin/ld_prague_theirs_boxed.jpg} \\
        (a) LucidDreamer (ZoeDepth~\cite{bhat2023zoedepth}) \\
        \includegraphics[width=\textwidth]{images/ld_plugin/ld_prague_ours.jpg} \\
        (b) LucidDreamer (with our depth completion model)\\
    \end{tabular}
    \caption{\textbf{Qualitative comparison of depth prediction backbones for LucidDreamer (Prague).} Consider the bridge discontinuities and ghosting artifacts, which disappear with our network.}
    \label{fig:ld-comparison-prague}
\end{figure}

\begin{figure}
    \centering
    \begin{tabular}{c}
        \includegraphics[width=\textwidth]{images/ld_plugin/ld_kyoto_theirs_boxed.jpg} \\
        (a) LucidDreamer (ZoeDepth~\cite{bhat2023zoedepth}) \\
        \includegraphics[width=\textwidth]{images/ld_plugin/ld_kyoto_ours.jpg} \\
        (b) LucidDreamer (with our depth completion model)\\
    \end{tabular}
    \caption{\textbf{Qualitative comparison of depth prediction backbones for LucidDreamer (Kyoto).} Our model leads to significantly less torn structures and provides an overall geometrically more sound scene. Please note that we manually removed visual obstructions from both scenes.}
    \label{fig:ld-comparison-kyoto}
\end{figure}

\begin{figure}
    \centering
    \begin{tabular}{c}
        \includegraphics[width=\textwidth]{images/ld_plugin/ld_nc_theirs_boxed.jpg} \\
        (a) LucidDreamer (ZoeDepth~\cite{bhat2023zoedepth}) \\
        \includegraphics[width=\textwidth]{images/ld_plugin/ld_nc_ours.jpg} \\
        (b) LucidDreamer (with our depth completion model)\\
    \end{tabular}
    \caption{\textbf{Qualitative comparison of depth prediction backbones for LucidDreamer (North Carolina).} With our depth completion network, the scene overall appears less distorted, in particular with regard to the foliage. Please note that we manually removed visual obstructions from both scenes.}
    \label{fig:ld-comparison-nc}
\end{figure}

\section{Discussion \& Limitations}
\label{sec:limitations}
In this section, we split our analysis of limitations into two distinct parts. First, we consider the limitations of our depth inpainting approach, specifically focusing on the trained network and the task of depth inpainting itself. Second, we provide a broader perspective on the limitations of the sequential scene generation task, which we present as an application of our depth completion model in line with other related works \cite{hollein23text2room:, zhang23text2nerf:, ouyang23text2immersion:, chung23luciddreamer:, yu23wonderjourney:, liu2021infinite}.

\paragraph{Depth Inpainting}
The performance of our depth inpainting model may be limited by a shift in the data distribution between training and inference. The data the model is trained on (such indoor scenes from NYU Depth or the datasets used to pre-train DPT~\cite{ranftl2021vision}, which forms the core of ZoeDepth~\cite{bhat2023zoedepth}), differ from the data it is typically applied to during 3D scene generation (\ie, synthetic images generated by Stable Diffusion). 
The differences may be due to the type of imagery (\eg, landscapes such as the \textit{Mountains in Peru} in \Cref{fig:supp-qualitative-results}), but also due to imperfections or artifacts caused by image generators that the depth network has never encountered during training.   

Another challenge lies in the inherently limited resolution of all depth estimation networks, which constrains the model's ability to accurately predict the depth of fine structures, especially around object boundaries.
As a consequence, these fine details might become detached from their corresponding objects during projection, which in turn affects the image inpainting step.
An example is shown in \Cref{fig:supp-fine-details}.

\paragraph{Scene Generation}
During scene generation, and similarly to prior work, we use supporting views to add further constraints to the Gaussian splat optimization process. 
How to optimally place the cameras for these views (to prevent ``crashing'' into the existing point cloud, see \Cref{fig:supp-supp-views}) remains an open question. 
More sophisticated camera pose generation algorithms, a strictly enforced minimum depth for points, or view-dependent translucency of parts of the point cloud might yield a more stable procedure to produce supporting views.

 \begin{figure}
     \centering
     \begin{tabular}{cc}
         \includegraphics[width=0.48\textwidth]{images/fine_details/dog_original.png} &
         \includegraphics[width=0.48\textwidth]{images/fine_details/dog_gs.png} \\
         Input image & \makecell{Projection based on depth prediction, \\ after depth snapping}
\end{tabular}
     \caption{\textbf{Loss of fine object details after projection.} Due to the limited resolution of depth predictions, fine details might be detached from their corresponding objects and become part of the background. We present an example of this for the given image, showing the resulting projection after applying our depth snapping to remove floating points (as outlined in Section 5.2). The fine hairs of the dog at its boundary have become part of the background.}
     \label{fig:supp-fine-details}
 \end{figure}

\begin{figure}
    \centering
    \begin{tabular}{cc}
        \includegraphics[width=0.46\textwidth]{images/supp_views/kyoto_supp_view_1.png} &
        \includegraphics[width=0.53\textwidth]{images/supp_views/kyoto_supp_view_2.png} \\
    \end{tabular}
    \caption{\textbf{Spurious points due to improperly placed supporting view cameras.} In this example, a supporting view camera was placed inside the point cloud, with some points obstructing its view. These points are rendered into an image that is used to guide the Gaussian splat optimization process, which cause them to be baked into the final Gaussian representation. These artifacts solely stem from our simplified method to place supporting view camera, as the offending generated frame had unexpectedly low depth values.}
    \label{fig:supp-supp-views}
\end{figure}

\clearpage
\bibliographystyle{splncs04}
\bibliography{main}